\documentclass[10pt, dvipsnames]{article} 
\usepackage[preprint]{tmlr}

\usepackage{commands_tmlr_math}
\usepackage{commands_math}
\usepackage{commands_table}

\usepackage{afterpage}
\usepackage{booktabs}
\usepackage{enumitem}
\usepackage{graphicx}
\usepackage{subcaption}
\usepackage{wrapfig}
\usepackage{algorithm}
\usepackage{algpseudocode}
\usepackage{marvosym}
\usepackage{hyperref}
\usepackage{url}

\graphicspath{{./figs/}}

\algnewcommand\Aand{\textbf{and }}
\algnewcommand\Or{\textbf{ or }}
\algnewcommand\Not{\textbf{not }}
\makeatletter 
 
\@addtoreset{algorithm}{chapter} 
\makeatother

\title{Enhanced gradient-based MCMC in discrete spaces}

\author{\name Benjamin Rhodes \email ben.rhodes@ed.ac.uk \\
      \addr University of Edinburgh
      \AND
      \name Michael Gutmann \email michael.gutmann@ed.ac.uk \\
      \addr University of Edinburgh
      }

\def\month{MM}  
\def\year{YYYY} 
\def\openreview{\url{https://openreview.net/forum?id=XXXX}} 

\begin{document}

\maketitle

\begin{abstract}
The recent introduction of gradient-based MCMC for discrete spaces holds great promise, and comes with the tantalising possibility of new discrete counterparts to celebrated continuous methods such as MALA and HMC. Towards this goal, we introduce several discrete Metropolis-Hastings samplers that are conceptually-inspired by MALA, and demonstrate their strong empirical performance across a range of challenging sampling problems in Bayesian inference and energy-based modelling. Methodologically, we identify why discrete analogues to \emph{preconditioned} MALA are generally intractable, motivating us to introduce a new kind of preconditioning based on auxiliary variables and the `Gaussian integral trick'.

\end{abstract}

\section{Introduction}

\begin{figure}[!b]
\centering
\includegraphics[width=.95\linewidth]{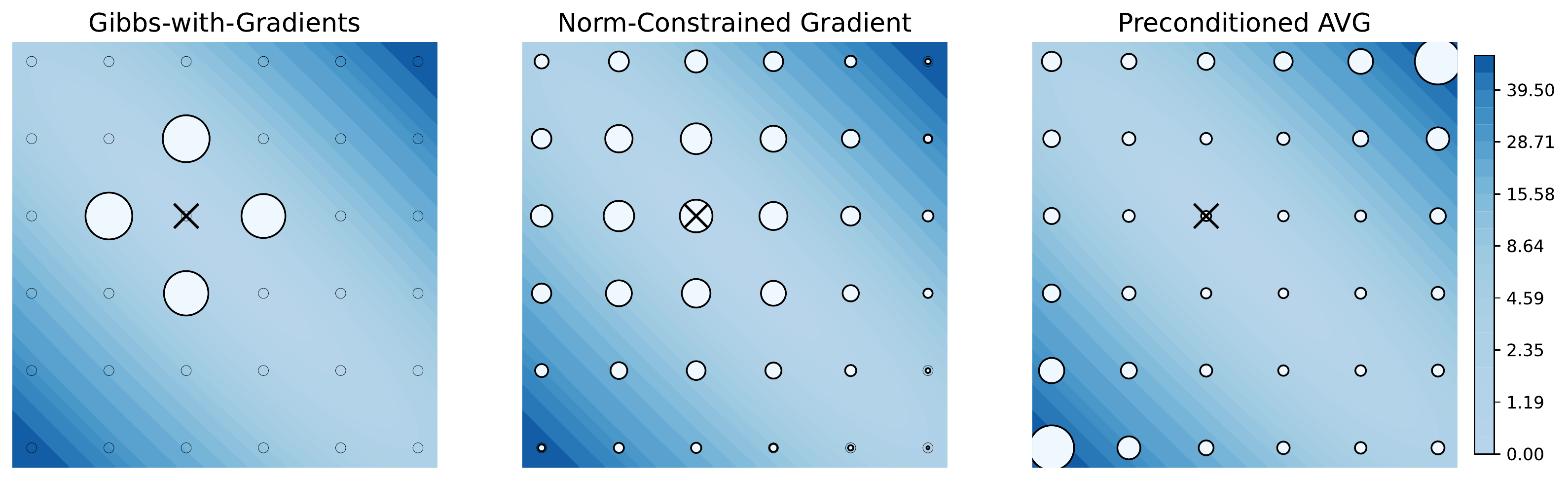}
 \caption{Metropolis-Hastings proposal distributions for an existing gradient-based sampler (left) and our proposed ones (centre \& right). The blue contours delineate a continuous function that, when restricted to the discrete lattice, equals the target log-probability function (up to a constant). The dots represent the proposal distribution given the current state of the Markov chain (black X).  The new samplers can update multiple dimensions at a time, which can lead to more efficient mixing of the Markov chains. In addition, the preconditioned sampler (right) accounts for the strong positive correlation between the two dimensions.}
\label{fig: overview of proposals}
\end{figure}

Gradient-based Markov Chain Monte Carlo (MCMC) offers an efficient, robust way to sample from a wide-class of probability distributions. The gradient serves as a concise descriptor of local geometry, which can be exploited when designing MCMC transition operators. In continuous spaces, such operators are often based on the Langevin diffusion \citep{roberts1996exponential, roberts1998optimal} or Hamiltonian Monte Carlo \citep{duane1987hybrid, neal2011mcmc}, which can be unified under a single complete framework \citep{ma2015complete}. Until recently, gradient-based operators were only viable for continuous distributions---after all, the standard gradient is undefined for discrete domains. However, \citet{grathwohl2021oops} made the important observation that many probability mass functions are naturally thought of as the \emph{restriction} of a continuous function (defined in e.g.\ $\mathbb{R}^d$) to a discrete subset (e.g.\ $\mathbb{N}^d$). Gradients in this ambient space can thus inform the design of a transition operator in the restricted discrete space. They demonstrate this via a promising new method, Gibbs-with-Gradients (GWG), that can be seen as a gradient-based version of the `locally informed proposals' introduced by \citet{zanella2020informed}.

Whilst the use of gradient-information makes GWG appealing, the method has multiple unresolved limitations. It:\vspace{0.5em}
\newline 
(L.1) cannot update more than one dimension at a time.
\newline
(L.2) uses a first-order Taylor expansion, which does not exploit second-order interactions in the target distribution.\vspace{0.5em}
\newline
In this paper, we will address these shortcomings via the introduction of several new discrete gradient-based samplers. Our strategy will be to first take a \emph{continuous} gradient-based sampler that does not have these limitations, namely the celebrated (preconditioned) Metropolis-Adjusted Langevin Algorithm (MALA), and use this as the conceptual basis for defining discrete analogues. This is not a straightforward task: MALA is typically viewed through the lens of stochastic differential equations (SDEs), which are not easily translated into discrete state-spaces. Fortunately, we identify two non-SDE characterisations of MALA that are more readily imported into the discrete setting, resulting in the following methods: 

\begin{itemize}
    \item The \textbf{Norm-Constrained Gradient (NCG)} sampler is constructed by viewing MALA as a locally-informed proposal \citep{zanella2020informed} whose domain is restricted to a discrete space. NCG essentially replaces the Hamming-ball constraint in Gibbs-with-Gradients with a soft norm constraint, enabling multiple dimensions to be updated at a time, thereby addressing limitation L.1. Unfortunately, preconditioning this sampler in a manner analogous to preconditioned MALA does not appear possible, leaving limitation L.2 unresolved.
    \item The \textbf{Auxiliary-Variable gradient (AVG)} sampler is constructed by viewing MALA as a marginalised auxiliary sampler \citep{titsias2018auxiliary} whose domain is restricted to a discrete space. The resulting sampler is conceptually and algebraically similar to NCG, but incurs extra computations that make it less appealing. However, AVG has the major advantage that it can be extended to include a kind of preconditioning, thereby addressing limitation L.2. Intriguingly, this \textbf{Preconditioned AVG (PAVG)} generalises an existing auxiliary-variable scheme \citep{martens2010parallelizable} that is only applicable to pairwise Markov Random fields (a.k.a.\ Boltzmann machines).
\end{itemize}

We believe the value of these new methods is two-fold: 1) Empirically, they show superior sampling efficiency compared to several important baselines across a range of problems. In particular, they outperform Gibbs-with-Gradients, validating our hypothesis that L.1 and L.2 are indeed limitations that should be addressed. 2) Methodologically, they demonstrate how new discrete gradient-based samplers can be derived via the frameworks of \citet{zanella2020informed} and \citet{titsias2018auxiliary}. To our knowledge, we are the first paper to use the latter framework for discrete problems, and our derivation of PAVG shows that auxiliary variables open up fundamentally new possibilities for constructing gradient-based samplers.

\section{Background}
\label{sec: background}
Our aim is to sample from distributions $p(\s)$ defined over a discrete domain $\mathcal{S}^d \subset \mathbb{R}^d$,
\begin{align}
    \label{eq: unnormalised model}
    \log p(\s) &= f(\s) - \log Z,& Z &= \sum_{\s \in \mathcal{S}^d} \exp(f(\s))
\end{align}
\vspace{-0.5cm}

where $Z$ is presumed unknown. For ease of exposition, we consider two important cases: binary vectors for which $\mathcal{S} = \{0, 1\}$ and finite ordinal data for which $\mathcal{S}=\{s^1, \ldots, s^k\}$ is an ordered set of increasing real numbers. We show in Appendix \ref{appendix: categorical vars} that our methods are also easily applicable to unordered categorical data represented as one-hot vectors. We assume $f$ is the restriction of a differentiable function defined over $\mathbb{R}^d$, and use $\nabla f$ to refer to its gradient. This assumption was first introduced by \citet{grathwohl2021oops} and it holds for many important distributions such as restricted Boltzmann machines, Ising models and deep energy-based models.

\subsection{Metropolis-Hastings}
A common approach to sampling such distributions is to use Metropolis-Hastings (MH) \cite{metropolis1953equation, hastings1970monte}, which evolves a chain of samples by iteratively sampling a proposal $\stt \sim q(\stt \given \st)$ and accepting with probability
\begin{align}
    \min \Big\{ 1,  \frac{\exp(f(\stt))}{\exp(f(\st))} \frac{q(\st \given \stt)}{q(\stt \given \st)} \Big\}.
\end{align}
The key challenge here is the choice of proposal. Roughly speaking, there are four desiderata \citep{robert1999monte} i) tractable to sample ii) tractable to evaluate iii) has reasonable acceptance rates and iv) the resulting chains have weak dependencies between successive states. Satisfying all four criteria is difficult, especially in high dimensions. Most `obvious' approaches fail to meet at least one criterion; for instance a uniform distribution over neighbouring states of $s_t$ can satisfy the first three criteria, but fail dramatically on the last.


\subsection{MALA and its preconditioned variant}
\label{sec: mala}
Metropolis-Adjusted Langevin Algorithm (MALA) \citep{roberts1996exponential, dwivedi2018log} is an effective \emph{continuous-space} MH sampler that uses gradient-based proposals of the form
\begin{align}
\label{eq: mala}
    q_{\epsilon}(\s \given \st) = \mathcal{N} \big( \s ; \st + \frac{\epsilon}{2} \nabla f(\st), \epsilon \mI \big).
\end{align}
where $\epsilon$ is a tunable step-size. If we were to skip the MH accept/reject step, then iteratively sampling this proposal corresponds to a discrete-time simulation of an SDE whose stationary distribution is the target $p(\s)$. Such time-discretisations induce errors which the MH step corrects for.

Many target distributions exhibit strong second-order interactions between dimensions. A natural way to handle this is via a suitable linear change of coordinates; a technique known as \emph{preconditioning}. Preconditioned MALA has proposals of the form
\begin{align}
\label{eq: pmala}
        q_{\epsilon}(\s \given \st) = \mathcal{N} \big( \s ; \st + \frac{\epsilon}{2} \Sigma \nabla f(\st), \epsilon \Sigma \big),
\end{align}
where $\Sigma$ is a user-specified symmetric semi-positive-definite matrix. This proposal also corresponds to a discretised SDE whose stationary distribution is $p(\s)$.

\subsection{Locally-informed proposals}
\citet{zanella2020informed} proposed a different framework for incorporating local geometry into an MH proposal, with the advantage of being applicable to discrete spaces. They define \emph{pointwise-informed proposals} of the form
\begin{align}
    q(\stt \given \st) = \frac{g(\exp(f(\stt)-f(\st))) K_{\sigma}(\stt \given \st)}{Z_g(\st)}
\end{align}
where $g$ is a non-negative `balancing' function, $K_{\sigma}$ is a symmetric kernel whose `width' is controlled by $\sigma$ (e.g.\ a uniform distribution over a local ball of radius $\sigma$) and $Z_g(\st)$ is a normalising constant.

The intuition here is that we take an `uninformed' kernel and re-weight it according to the target distribution, biasing our proposal in favour of higher density regions. \citet{zanella2020informed} identify a class of balancing functions $g$ that are optimal when the kernel is sufficiently local (i.e.\ $\sigma$ is small); an important member of that class is the square-root function $g(x)=\sqrt{x}$.

\subsection{Gibbs-with-Gradients}
A major challenge when using locally-informed schemes in discrete spaces is the cost of normalising and sampling the proposal distribution. Even for the most simple choice of kernel---a uniform distribution over a Hamming ball of radius 1---the cost is $\mathcal{O}(d)$ evaluations of $f$ for a $d$-dimensional problem. To reduce this cost, \citet{grathwohl2021oops} use the innovative trick of treating $f$ as a continuous function and approximating it with a first-order Taylor expansion about the current state $\st$, giving the `first-order local proposal'
\begin{align}
\label{eq: gwg local-proposal}
    q(\rvs \given \st) =  \frac{ \exp \Big( \frac{1}{2} \nabla f(\st)^T (\rvs - \st) \Big) \mathbb{I}(\rvs \in H_1(\st))}{
    \sum_{\rvs \in H_1(\st)} \exp \Big( \frac{1}{2} \nabla f(\st)^T (\rvs - \st) \Big)}
\end{align}
where $H_1(\st)$ is a Hamming ball of radius 1 around $\st$. The normalising constant of this distribution is cheap to compute: a single gradient computation and a sum over $d$ inexpensive terms. 

Unfortunately, there are two significant limitations with this proposal: i) updating \emph{one} dimension at a time can be slow, and a straightforward approach to `broaden' this proposal by using a larger radius is prohibitively expensive due to the rapidly increasing size of the Hamming ball ii) linear approximations of the target distribution combined with symmetric kernels cannot account for second-order interactions between dimensions.




\section{Discrete analogues to MALA}
Our goal in this section is to design new discrete gradient-based MCMC schemes that overcome the aforementioned limitations. We achieve this by `importing' two different characterisations of the Metropolis-adjusted Langevin Algorithm (MALA) into a discrete setting.

\subsection{Norm-constrained gradient sampler: NCG}
As discussed, the use of Hamming balls in Gibbs-with-Gradients makes it challenging to update more than one dimension at a time. Instead, we propose to use a soft norm `constraint', transforming \eqref{eq: gwg local-proposal} into
\begin{align}
\label{eq: ncg local-proposal}
    q_{\epsilon}(\rvs \given \st) \propto \exp \Big( \frac{1}{2} \nabla f(\st)^T (\rvs - \st) \Big) \exp \Big( -\frac{1}{2\epsilon} \norm{\rvs - \st}_2^2 \Big).
\end{align}
Such a first-order informed proposal was already discussed by \citet{zanella2020informed} in the context of \emph{continuous spaces}, where, after normalising, it corresponds exactly to the proposal distribution used by Metropolis-adjusted Langevin Algorithm (MALA) in \eqref{eq: mala}.

If we instead restrict $\rvs$ to a discrete state-space $ \mathcal{S}^d \subset \mathbb{R}^d$, we obtain the fully factorised distribution\footnote{The definition of $\sigma(\ervx)$ abuses notation by assuming $\ervx$ is a function of $\ervs \in \mathcal{S}$, and that we sum over all values of $\ervs$ in the denominator.}
\begin{align}
\label{eq: ncg local-proposal factorised} q_{\epsilon}(\rvs \given \st) &= \prod_{i=1}^d \sigma \Big(\Big[ \frac{1}{2} \nabla f(\st)_i + \frac{1}{\epsilon} \sti \Big] \si - \frac{1}{2\epsilon} \si^2 \Big), & \sigma(\ervx) := \frac{\exp(\ervx)}{\sum_{\mathcal{S}} \exp(\ervx)} 
\end{align}
Due its factorised structure, this proposal is efficient to evaluate and sample, making it straightforward to use in a Metropolis-Hastings scheme; see Algorithm \ref{algo: NCG} \footnote{The NCG method was independently discovered by \citep{zhang2022langevin}, see related work discussion in Section \ref{sec: related work}}. Like MALA, it has a step-size parameter $\epsilon$ that controls the `width' of the distribution and larger step-sizes enable proposals $\stt$ that differ from $\st$ in multiple dimensions. We note that NCG should be viewed as an alternative to (and not a generalisation of) GWG, since there is no setting of the step-size $\epsilon$ that recovers GWG.

\subsubsection{The problem of preconditioning}
\label{sec: the problem of preconditioning}
A natural way to precondition NCG would be to cast PMALA (\eqref{eq: pmala}) as a first-order informed proposal and then restrict the domain to a discrete subspace. Unfortunately, we find that this approach is not viable. To see why, we first note that by replacing the Euclidean-norm $\norm{\s - \st}^2$ in \eqref{eq: ncg local-proposal} with a Mahalanobis-norm $(\s - \st)^T \Sigma^{-1} (\s - \st)$, we obtain a proposal distribution that, for continuous $\s \in \mathbb{R}^d$, corresponds exactly to PMALA. However, when we restrict $\s$ to a discrete domain, this distribution becomes a pairwise Markov random field that, in general, is intractable to normalise and sample from, making it unusable as an MH proposal. 

One could avoid this intractability by choosing a highly restricted, sparse form for $\Sigma^{-1}$ (e.g.\ diagonal); we leave this direction to future work, and instead focus on the more general problem of constructing a gradient-based sampler that can work with arbitrary $\Sigma$. To solve this challenge, we depart from the locally-informed framework of \citet{zanella2020informed} and introduce an alternative MH-framework that operates in an extended state-space. This framework also contains a discrete analogue to MALA that resembles NCG, with the key distinction that it admits a kind of tractable preconditioning.

\subsection{(Preconditioned) Auxiliary Variable Gradient sampler: AVG and PAVG}
\label{sec: avg and pavg}

As explained by \citet{titsias2018auxiliary}, MALA can be derived as an auxiliary variable scheme. To see this, we first augment our continuous state $\s \in \mathbb{R}^d$ with Gaussian auxiliary variables $\z \in \mathbb{R}^d$, such that our unnormalised target density becomes $ \pi(\s, \z) = \exp(f(\s)) \mathcal{N}(\z; \s / \sqrt{\epsilon/2}, \mI)$. In theory, this distribution could be sampled in a block-Gibbs fashion, by first sampling $\zt \sim \mathcal{N}(\z; \st / \sqrt{\epsilon/2}, \mI)$, and then sampling $\stt \sim \pi(\s \given \zt) \propto \pi(\s, \zt)$. However, for general functions $f$ this second sampling step is intractable, so we replace it with an MH accept-reject step using the proposal distribution:
\begin{align}
    q_{\epsilon}(\s \given \zt, \st)  &\propto \exp(f(\st) + \nabla f(\st)^T (\s - \st)) \mathcal{N}(\zt; \s / \sqrt{\epsilon/2}, \mI) \label{eq: avg unnormalised proposal} \\
    & = \mathcal{N}(\s; \sqrt{\epsilon/2} \zt + (\epsilon/2)  \nabla f(\st), (\epsilon/2) \mI),
\end{align}
where \eqref{eq: avg unnormalised proposal} approximates $\pi(\s, \zt)$ via a Taylor expansion. Equipped with this proposal, we can now perform block-wise MH sampling in the extended state space. However, we could also chose to marginalise out the latent variables, and doing so yields the MALA proposal
\begin{align}
q_{\epsilon}(\s \given \st) = \int N(\z; \st/\sqrt{\epsilon/2}, \mI) q_{\epsilon}(\s \given \zt, \st) d\z = \mathcal{N}(\s; \st + (\epsilon/2)  \nabla f(\st), \epsilon \mI).
\label{eq: marginalised auxiliary MALA}
\end{align}
The auxiliary variable procedure just described is equally applicable to discrete state-spaces, using the discrete Taylor approximation `trick' of \citet{grathwohl2021oops}. If we take $\s$ to belong to a discrete state-space $ \mathcal{S}^d \subset \mathbb{R}^d$ (but continue to use Gaussian auxiliary variables), then the proposal $q_{\epsilon}(\s \given \zt, \st)$ in \eqref{eq: avg unnormalised proposal} can be written as a fully-factorised distribution $\prod_{i=1}^d  q_{\epsilon, i}(\si \given \zti, \st)$ where each factor has the form
\begin{align}
\label{eq: avg univariate factor}
q_{\epsilon, i}(\si \given \zti, \st) = 
\sigma \Big(\Big[ \nabla f(\st)_i + \sqrt{\frac{2}{\epsilon}} \zti  \Big] s_i - \frac{1}{\epsilon} s_i^2 \Big),  &
 & \sigma(\ervx) = \frac{\exp(\ervx)}{\sum_{\mathcal{S}} \exp(\ervx)}.
\end{align}
We can easily evaluate and sample this proposal distribution, enabling us to perform block-wise MH sampling as summarised in Algorithm \ref{algo: AVG}. We refer to this method as the Auxiliary Variable Gradient (AVG) sampler.

Instead of using block-wise MH, one might try to marginalise out $\z_t$, giving $q_{\epsilon}(\s \given \st) =  \prod_{i=1}^d  \mathbb{E}_{\zti} \Big[ q_{\epsilon, i}(\si \given \zti, \st) \Big]$.
However, we do not use such a marginalised proposal in this paper, since the required expectations have no closed-form solution, necessitating careful numerical approximations. More fundamentally, we focus on the block-wise scheme as it admits an effective kind of preconditioning.

\subsubsection{The promise of preconditioning}
\label{sec: pavg}

A straightforward attempt to discretise PMALA  via auxiliary-variables runs into difficulties. To obtain the PMALA proposal in  \eqref{eq: pmala}, we need to replace the conditional Gaussian in \eqref{eq: avg unnormalised proposal} by $\mathcal{N}(\z_t; (\epsilon/2 \Sigma)^{-1/2} \s, \mI)$. Unfortunately, the corresponding discrete proposal distribution is an intractable pairwise Markov random field---see Appendix \ref{appendix: aux var precon mala}, for details.

However, we now show that a different kind of `preconditioning' is viable. We start by making a second-order approximation of the target
\begin{align}
\label{eq: global quad approx}
    f(\s) \approx f(\st) + \nabla f(\st)^T (\s - \st) + (1/2)(\s - \st)^T \Sigma (\s - \st),
\end{align}
where the second-order term uses a \emph{global} (independent of $t$) matrix $\Sigma$. The nature of this approximation, and how to choose $\Sigma$, will be discussed in the next section. 

We then define an unnormalised joint distribution  $ \pi(\s, \z) = \exp(f(\s)) \ \mathcal{N}(\z; \Sigeps^{1/2}\st, \mI)$, where $\Sigeps := \Sigma + (2/\epsilon) \mI$. Just like in our derivation of AVG, block-Gibbs sampling of $\z$ and $\s$ is prevented by the intractable $\pi(\s \given \z_t) \propto \pi(\s, \zt)$. So we use an MH step, approximating $\pi(\s, \zt)$ to obtain the proposal
\begin{align}
    q_{\epsilon}(\s \given \zt, \st) &\propto \exp(f(\st) + \nabla f(\st)^T (\s - \st) + (1/2)(\s - \st)^T \Sigma (\s - \st)) \ \mathcal{N}(\z; \Sigeps^{1/2}\s, \mI) \label{eq: pavg auxiliary proposal 1} \\
    &\propto \exp \big( \big[\nabla f(\st) - \Sigma \st + \Sigeps^{1/2} \zt \big]^T \s - (1/\epsilon) \s^T\s \big). \label{eq: pavg auxiliary proposal 2}\\
    &= \prod_{i=1}^d \sigma\Big( \Big[ \nabla f(\st)_i -  (\Sigma \st)_i + (\Sigeps^{1/2} \zt)_i   \Big] \si - \frac{1}{\epsilon} \si^2 \Big), \hspace{10mm} \text{where} \hspace{3mm} \sigma(\ervx) = \frac{\exp(\ervx)}{\sum_{\mathcal{S}} \exp(\ervx)}, \label{eq: pavg auxiliary proposal 3}
\end{align}
which is fully-factorised and thus tractable to evaluate and sample. The resulting block-wise MH sampling scheme is called Preconditioned AVG (PAVG) and is summarised in Algorithm \ref{algo: PAVG}. By comparing \eqref{eq: avg univariate factor} and \eqref{eq: pavg auxiliary proposal 3}, we see that setting $\Sigma = 0$ recovers AVG.
 
 
 The key step in obtaining a factorised proposal occurs when going from \eqref{eq: pavg auxiliary proposal 1} to \eqref{eq: pavg auxiliary proposal 2}, since the quadratic terms in $\s$ cancel out. Using Gaussian auxiliary variables to induce such cancellations actually has a long history in statistical physics where it is known as the Hubbard-Stratonovich transform \citep{hubbard1959calculation} or the "Gaussian integral trick" \citep{hertz1991introduction}. The trick has also been used in the Machine learning literature; first by \citet{martens2010parallelizable} and then extended by \citet{zhang2012continuous}. In both these prior works however, the target function was exactly quadratic i.e.\ $f(\s) = \rvb^T\s + \s^T \Sigma \s$, in which case the proposal distribution above is `exact' and an MH accept/reject step is unnecessary. Thus, our proposed approach, which we refer to as PAVG, can be seen as an extension of \citet{martens2010parallelizable} to more general target distributions.\footnote{Important caveat: \citet{martens2010parallelizable} allow $\Sigma$ to have negative eigenvalues. We can also allow this, with small modifications to our proposal distributions as described in Appendix \ref{appendix: choice of precon mat}}
 
\subsubsection{Choice of matrix $\Sigma$}
\label{sec: choice of precon mat}
The above derivation assumed a kind of quadratic approximation of $f(\s)$ in \eqref{eq: global quad approx}. We may hope that such an approximation is reasonable whenever there are global second-order interactions in the target distribution. We propose to capture such interactions by building on ideas from adaptive continuous-space MCMC \citep{rosenthal2011optimal}, where one can, for instance, estimate an empirical covariance/precision matrix $\Sigma_{emp}$ from a `dataset' of samples $\mathcal{D}$ obtained during an initial burn-in period. Given this matrix, we then use a common procedure (e.g.\ see Algorithm 4 of \citet{andrieu2008tutorial}) of re-scaling it by an adaptively learned parameter $\gamma$. We give full details of our adaptive procedure in Appendix \ref{appendix: choice of precon mat}. A benefit of this re-scaling is that we can automatically `fallback' to AVG by learning $\gamma = 0$. However, in practice, we often find that values of $\gamma$ far from $0$ or $1$ are learned.


\section{Related Work}
\label{sec: related work}

Most closely related to our work is the concurrent paper of \citet{zhang2022langevin}, who introduce a discrete `langevin-like' sampler called DMALA that coincides with NCG. Beyond this shared method, our two works diverge in multiple ways. \citet{zhang2022langevin} explore variants of DMALA that drop the Metropolis accept-reject step, or use \emph{diagonal} preconditioning; a possibility we briefly discussed in Section \ref{sec: the problem of preconditioning}. In contrast, our work is the first to identify the intractability of general preconditioning matrices within the DMALA/NCG framework, which motivated our novel MALA-inspired auxiliary variable schemes.

A range of prior works have `mapped' discrete spaces into continuous ones, thereby enabling MALA/HMC in the continuous space. 
These methods are generally specialised to particular forms of target distribution \citep{zhang2012continuous}, or to certain data types e.g.\ binary data \citep{pakman2013auxiliary} or trees \citep{dinh2017probabilistic}. More problematically, the embedded distributions are typically piece-wise continuous \citep{nishimura2017discontinuous} and highly multi-modal, which may explain their limited empirical success \citep{zanella2020informed, grathwohl2021oops}. 

\section{Experiments}
\label{sec: experiments}

We evaluate the newly proposed methods---NCG, AVG \& PAVG--- on four problem types: 1) sampling from highly correlated ordinal mixture distributions 2) a sparse Bayesian variable selection problem 3) estimation of Ising models and 4) sampling a deep energy-based model parameterised by a convolutional neural network.

Our key baselines are Gibbs-with-Gradients (GWG) and a standard Gibbs sampler \citep{geman1984stochastic}. For our ordinal experiments, we also compare to a Metropolis-Hastings sampler with uniform proposal over a local ball  of radius $r$ and a simple effective extension of GWG that we introduce: `ordinal-GWG'. In Appendix \ref{appendix: baselines}, we provide details on these baselines, including how we tune step-size parameters to maximise $\norm{\stt - \st}_1$ between successive states of the MCMC chains, similar to e.g.\ \citep{levy2018generalizing}. 

\begin{figure}[!t]
\centering
\begin{subfigure}{0.32\textwidth}
  \centering
  \includegraphics[width=.99\linewidth]{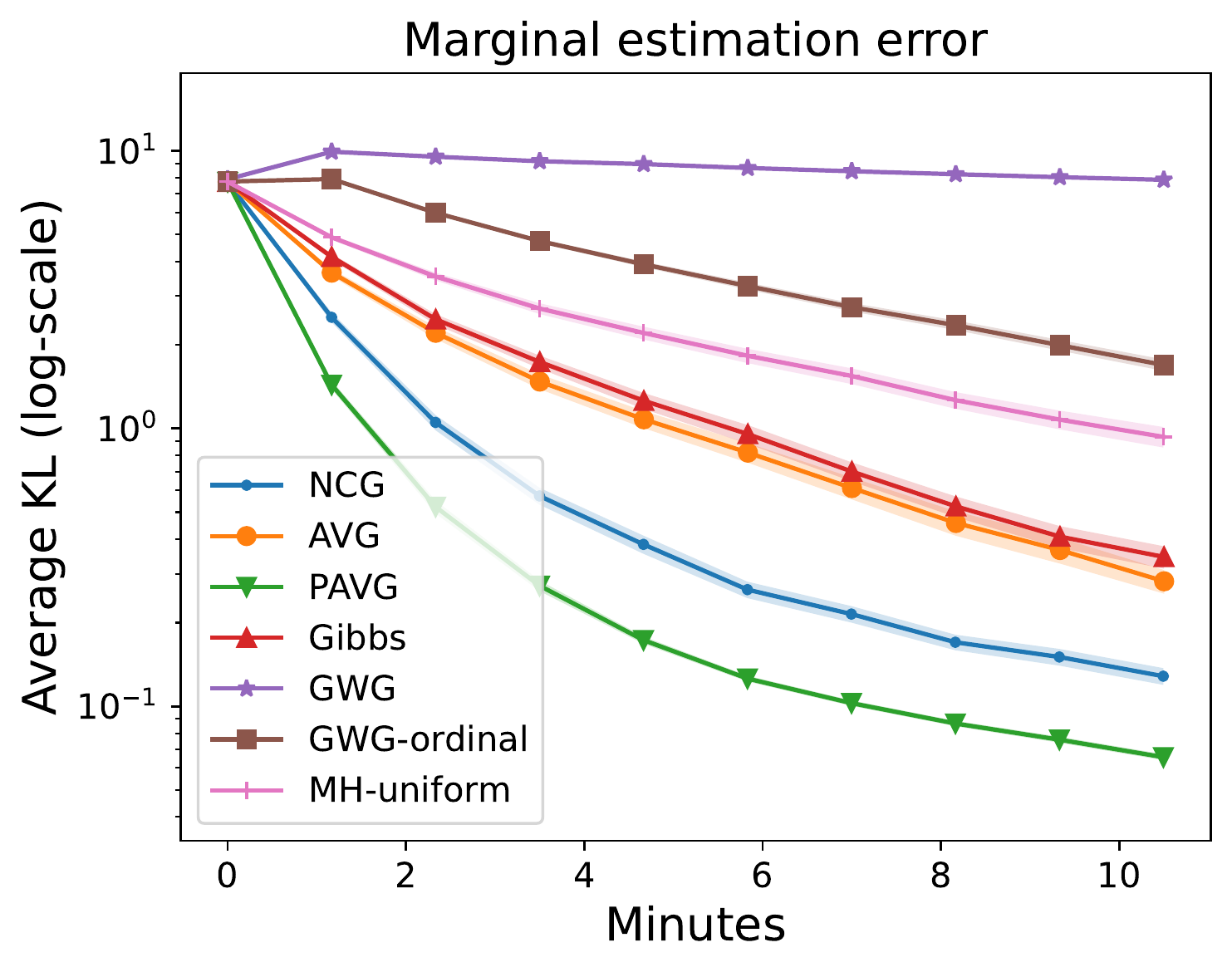}
\end{subfigure}\hfill
\begin{subfigure}{0.32\textwidth}
  \centering
  \includegraphics[width=.99\linewidth]{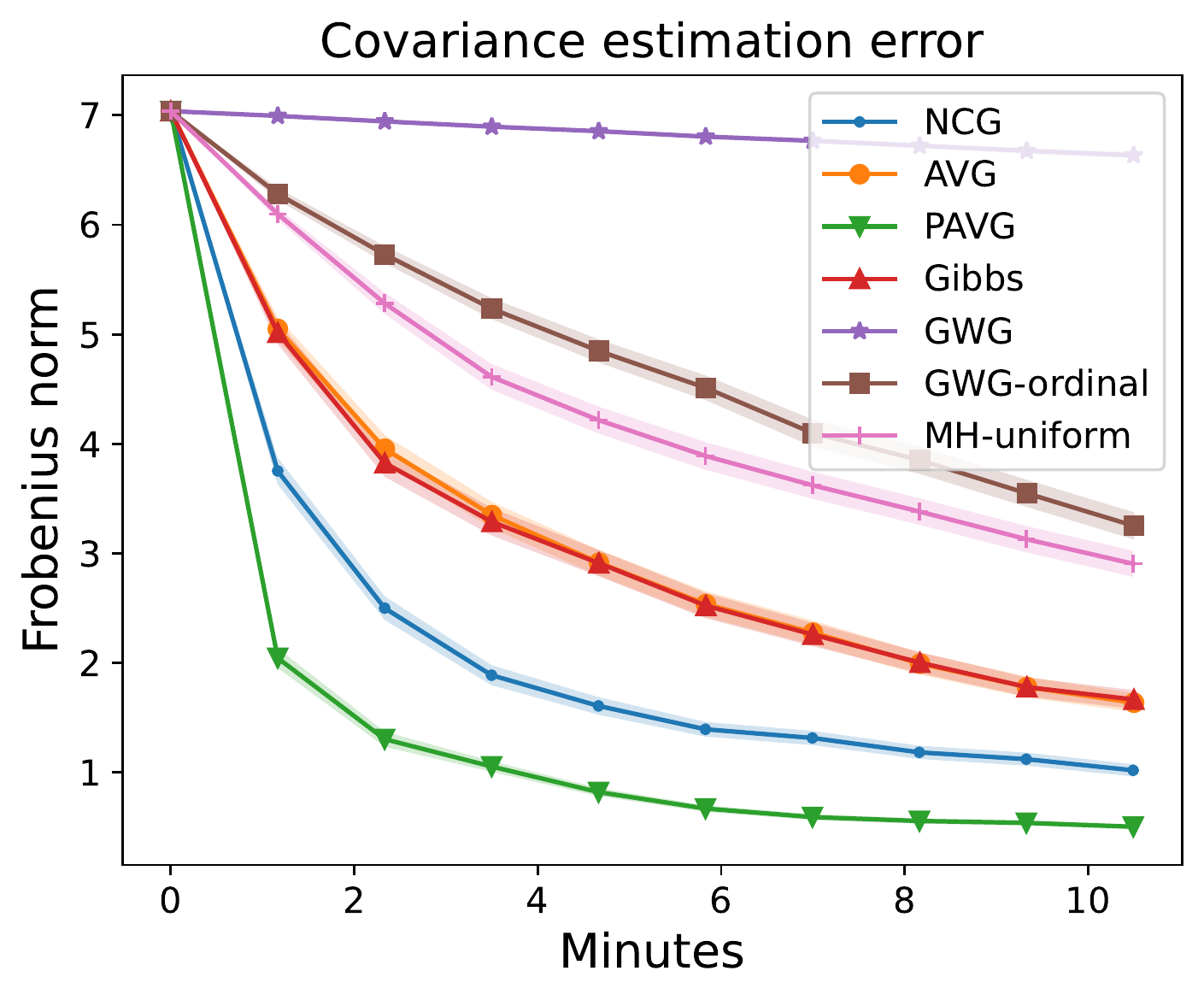}
\end{subfigure}\hfill
\begin{subfigure}{0.32\textwidth}
  \centering
  \includegraphics[width=.99\linewidth]{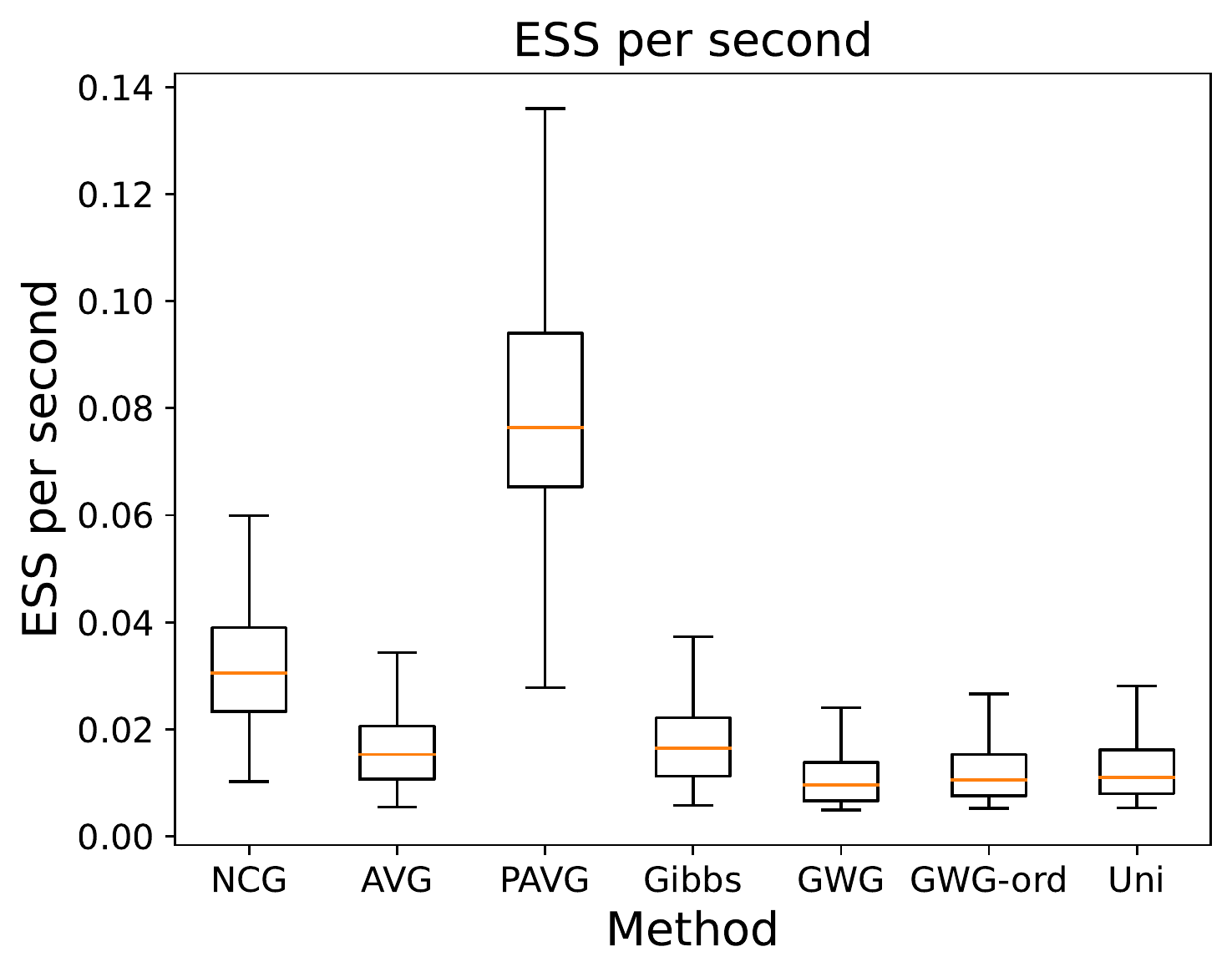}
\end{subfigure}
\begin{subfigure}{0.32\textwidth}%
  \centering
  \includegraphics[width=.99\linewidth]{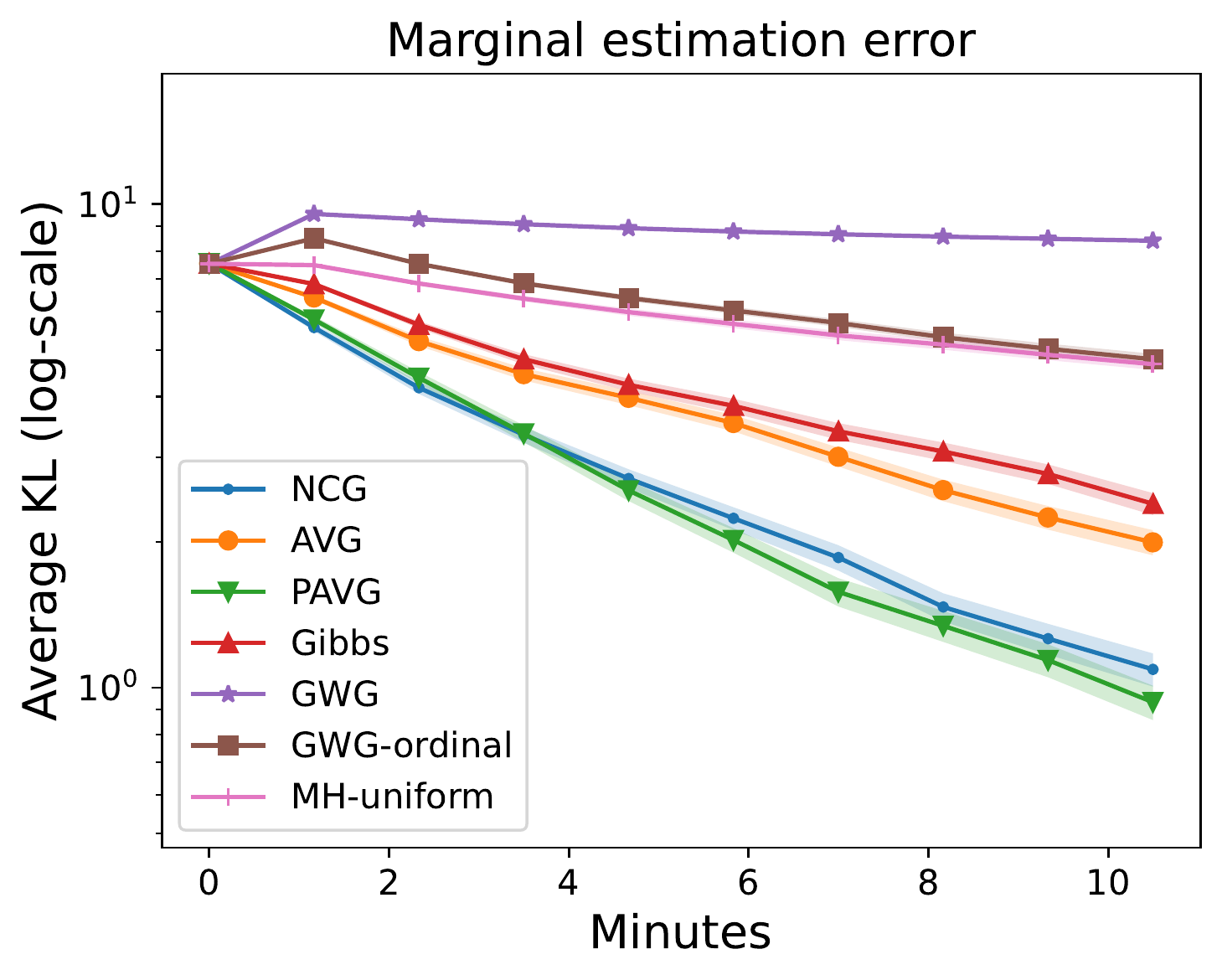}
\end{subfigure}\hfill
\begin{subfigure}{0.32\textwidth}
  \centering
  \includegraphics[width=.99\linewidth]{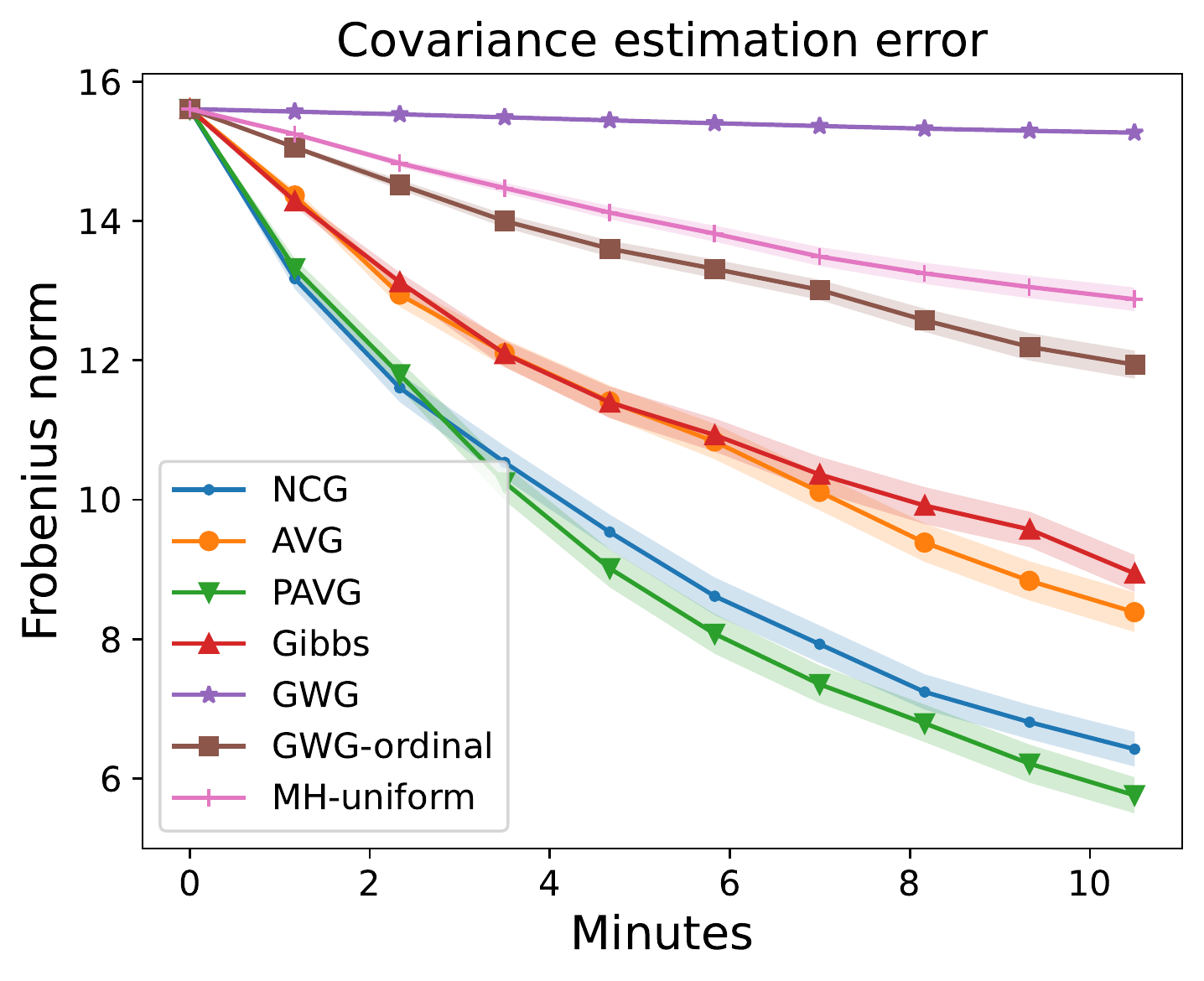}
\end{subfigure}\hfill
\begin{subfigure}{0.32\textwidth}%
  \centering
  \includegraphics[width=.99\linewidth]{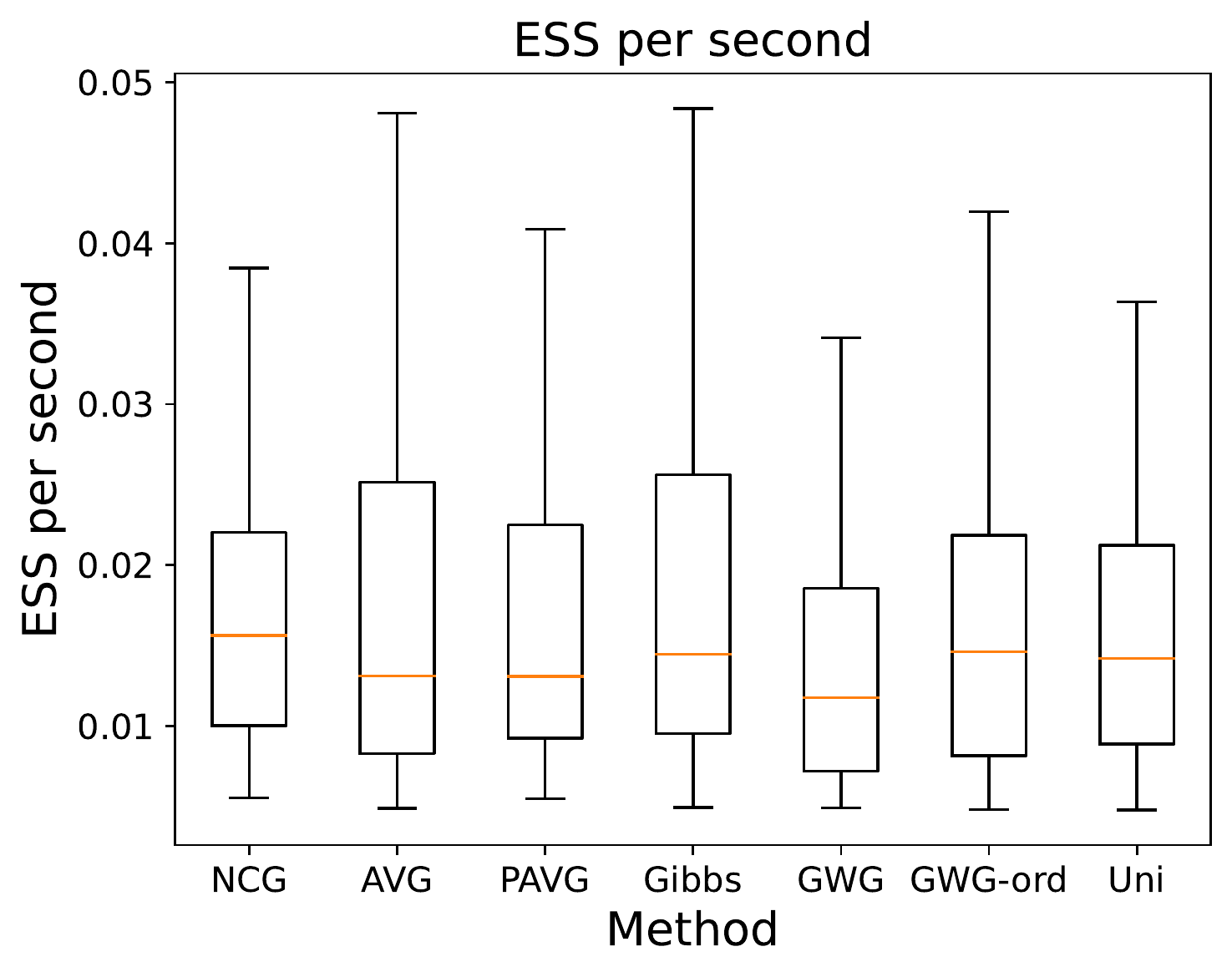}
\end{subfigure}
 \caption{20D mixture-of-polynomial results.
 Top row: results for 2\textsuperscript{nd} order polynomial in \eqref{eq: 2nd order poly}.
 Bottom row: results for 4\textsuperscript{th} order polynomial in \eqref{eq: 4th order poly}.
Left column: KL divergence between true and estimated marginals, averaged over all dimensions. Middle: Estimation error of the empirical covariance matrix. Right: Effective sample size per second; higher is better. All error bars computed across 100 parallel chains.}
  \label{fig: ordinal results}
\end{figure}

\newpage
\subsection{20D ordinal mixture-of-polynomials}
\label{sec: ordinal experiments}

\begingroup
\setlength{\columnsep}{1em}%
\begin{wrapfigure}{R}{0.17\textwidth} 
\vspace{-3.5em}
\begin{subfigure}{0.17\textwidth}
\centering
\includegraphics[width=.99\linewidth]{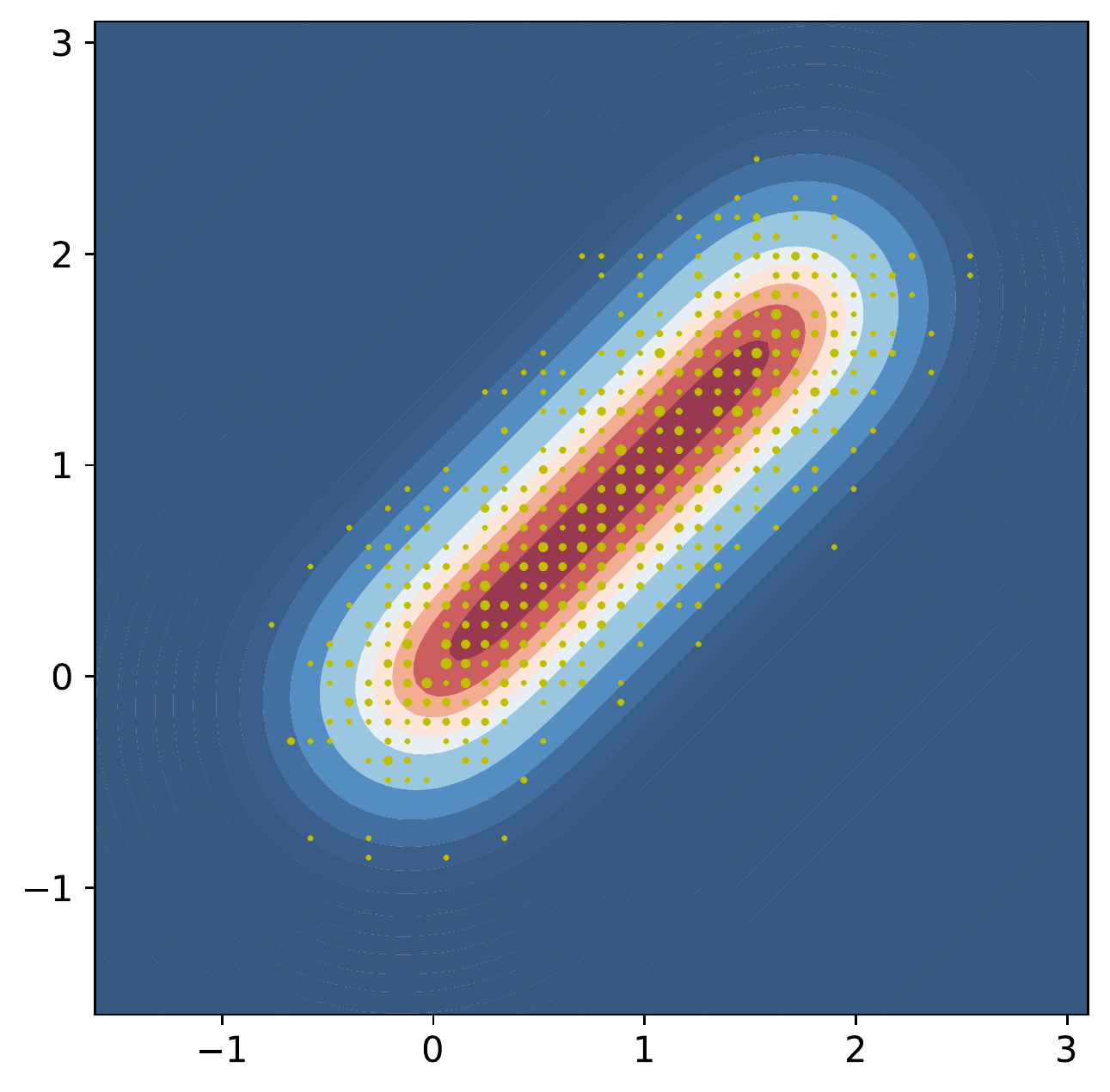}
\end{subfigure}
\begin{subfigure}{0.17\textwidth}%
\centering
\includegraphics[width=.99\linewidth]{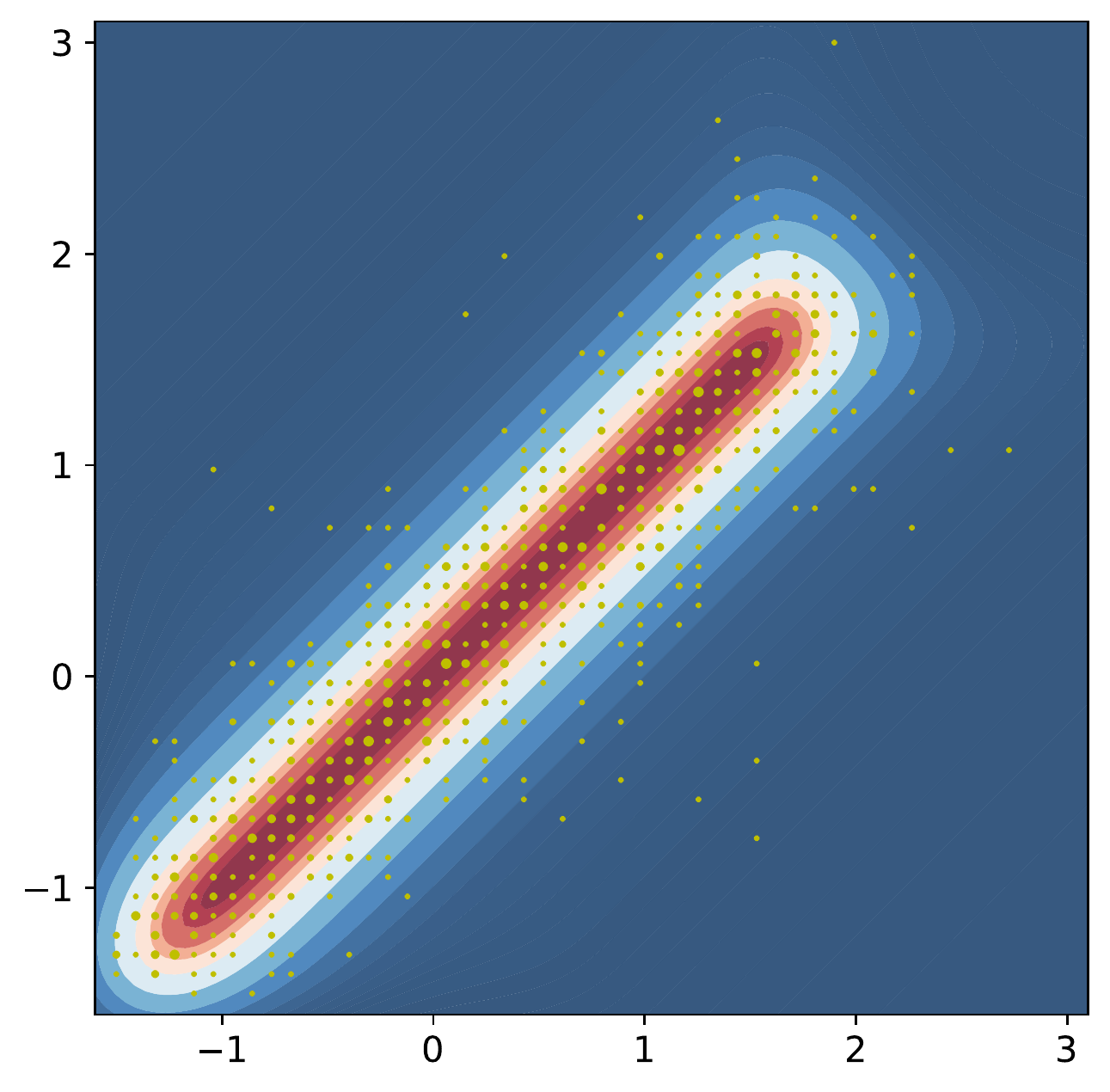}
\end{subfigure}
\caption{\small 2D ordinal illustrations. \eqref{eq: 2nd order poly} \& \eqref{eq: 4th order poly} correspond to top \& bottom, respectively.}
\label{fig: 2d ordinal dists}
\end{wrapfigure}

We define highly-correlated ordinal target distributions $\log p(\s) = \exp(f(\s)) - \log Z$ over $20$-dimensional lattices $\mathcal{S}^{20} \subset \mathbb{R}^{20}$, where $\mathcal{S}$ contains 50 equally spaced points in the interval $[-1.5, 3.0]$. We construct these target distributions from mixtures of fully-factorised distributions, enabling exact sampling and evaluation of the normaliser $Z$. $f$ has the form
\begin{align}
\label{eq: ordinal mixture formula}
\vspace{-1em}
    f(\s) &= \log \Big( \sum_{k=1}^{50} \exp \Big( \sum_{i=1}^{20} g_k(\si) \Big) \Big)
\end{align}
where $k$ indexes a component of the mixture distribution, and $g_k: \mathcal{S} \rightarrow \mathbb{R}$ is a polynomial that we allow to take one of two forms:
\begin{align}
     \textbf{(2\textsuperscript{nd} order)} \hspace{2em} g_k(\ervu) &= 1.5 - 2\ervt_k - 6\ervt_k^2,  \hspace{4em} \ervt_k := \ervu+ k/25 \label{eq: 2nd order poly} \\
     \textbf{(4\textsuperscript{th} order)} \hspace{2em} g_k(\ervu) &= -\ervt_k + \ervt_k^2 - \ervt_k^3 - \ervt_k^4, \hspace{2em} \ervt_k := (2\ervu-1+ 3k/50). \label{eq: 4th order poly}
\end{align}

We visualise 2D versions of the resulting target distributions in Figure \ref{fig: 2d ordinal dists}. The two dimensions are highly correlated with probability mass concentrating along the main diagonal. Similarly, in 20 dimensions, mass concentrates along the diagonal of a hypercube, meaning \emph{all} dimensions are positively correlated.

\endgroup

We track the similarity between the empirical distribution of MCMC samples $q(\s)$ and the target distribution $p(\s)$ using the following two metrics i) marginal estimation error: the average KL-divergence between marginals $(1/d) \sum_{i=1}^d \infdiv{KL}{q_i}{p_i}$ and ii) covariance estimation error: the difference, in Frobenius norm, between the empirical covariance matrices estimated under both distributions. We also track the Effective Sample Size (ESS) of each sampler. Full details of these evaluation metrics in Appendix \ref{appendix: ordinal and sbl eval metrics}.

\subsubsection{Results}

Figure \ref{fig: ordinal results} shows the results, with the top \& bottom rows corresponding to the 2\textsuperscript{nd} \& 4\textsuperscript{th} order polynomials, respectively. In the 2\textsuperscript{nd} order case, the evaluation metrics imply the following ranking:
\begin{align}
    \text{PAVG > NCG > AVG = Gibbs > MH-uniform > GWG-ordinal > GWG}
\end{align}
This ranking shows that all newly proposed samplers are either competitive or superior to baselines, with the preconditioned sampler, PAVG, showing especially strong performance. It is slightly surprising that GWG-based methods would perform \emph{worse} than a standard Gibbs sampler, however this is partly due to higher wall-clock costs; GWG-ordinal actually matches Gibbs \emph{per-iteration}.

For the 4\textsuperscript{th} order polynomial (bottom row), the estimation error metrics (columns 1 \& 2) imply a similar ranking to before. The main difference is that PAVG and NCG are now matched in performance. This relative reduction in the performance of PAVG is not entirely surprising: the derivation of PAVG relied on a kind of global quadratic approximation of $f(\s)$ (\eqref{eq: global quad approx}). Intuitively, it makes sense that our approximation degraded with the addition of 3\textsuperscript{rd} and 4\textsuperscript{th} order terms in the polynomial of \eqref{eq: 4th order poly}. 

Finally, we note that whilst the ESS estimates in the top row correlate well with the other two metrics, the ESS estimates in the bottom row are less concordant. We believe this may indicate sub-optimal estimation of ESS that can occur when the target distribution is heavy-tailed \citep{vehtari2021rank}.

\subsection{100D Sparse Bayesian linear regression}
\label{sec: sparse bayes experiment}

\begin{figure}[!t]
\centering
\begin{subfigure}{0.32\textwidth}
  \centering
  \includegraphics[width=.99\linewidth]{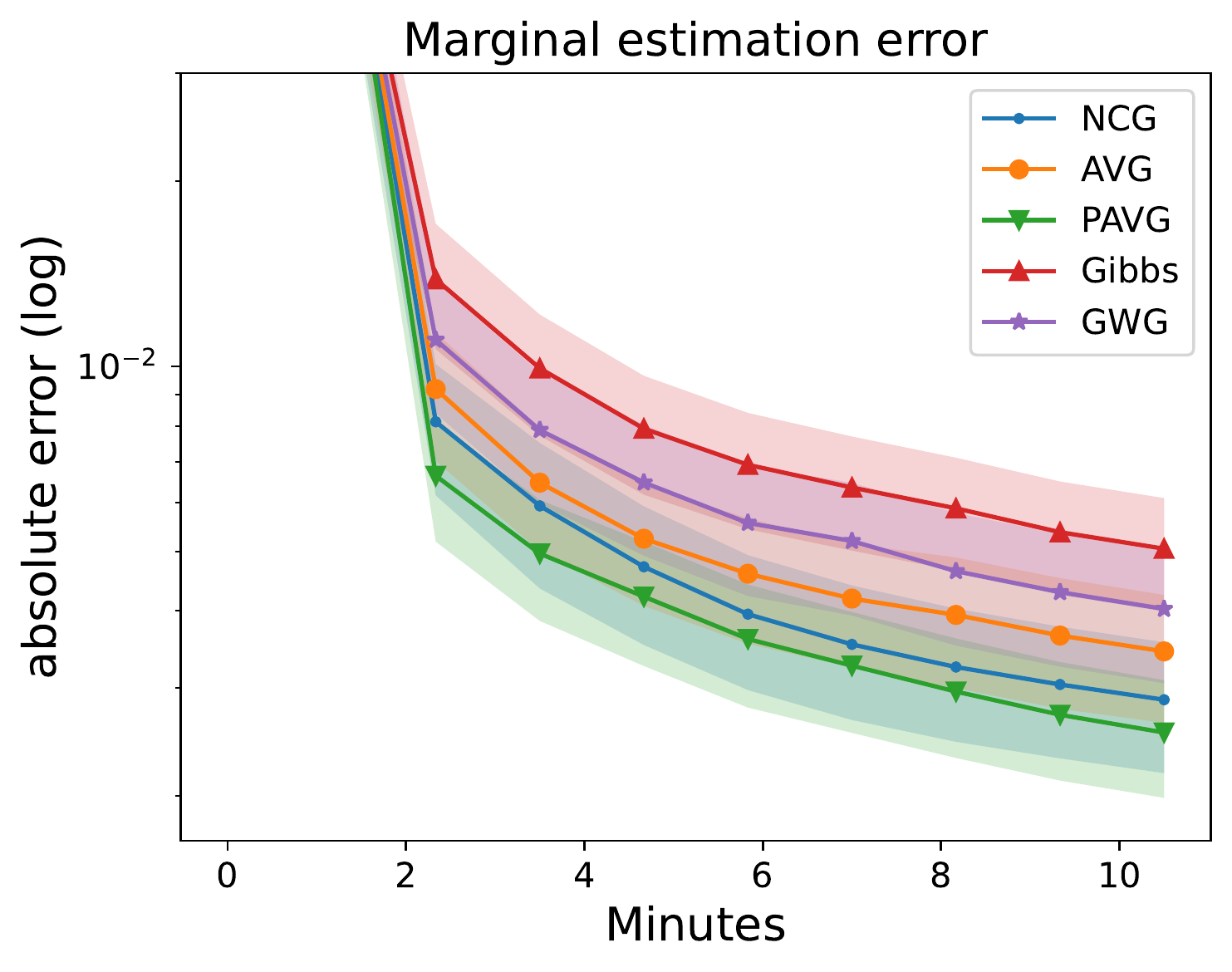}
\end{subfigure}\hfill
\begin{subfigure}{0.32\textwidth}
  \centering
  \includegraphics[width=.99\linewidth]{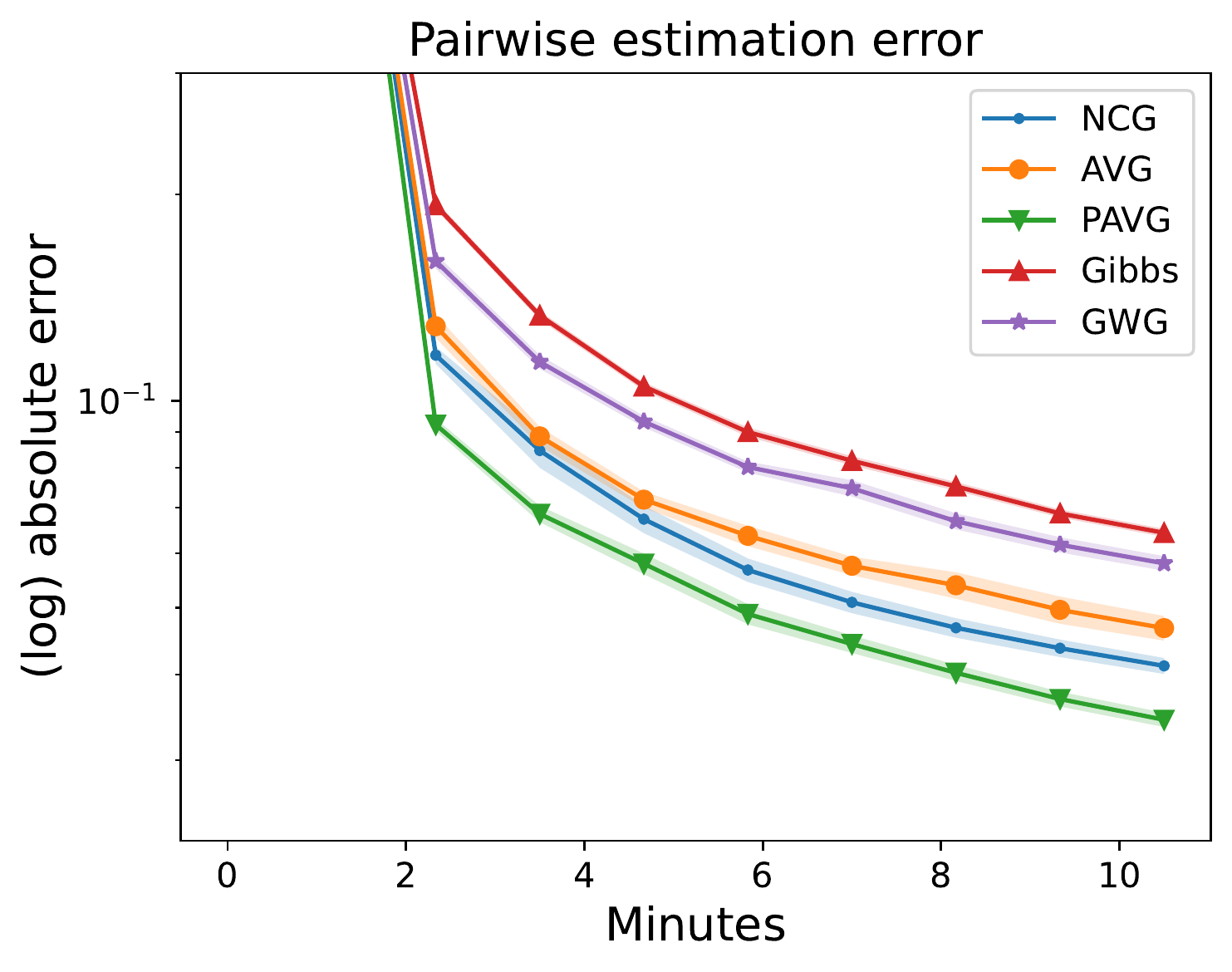}
\end{subfigure}\hfill
\begin{subfigure}{0.32\textwidth}
  \centering
  \includegraphics[width=.99\linewidth]{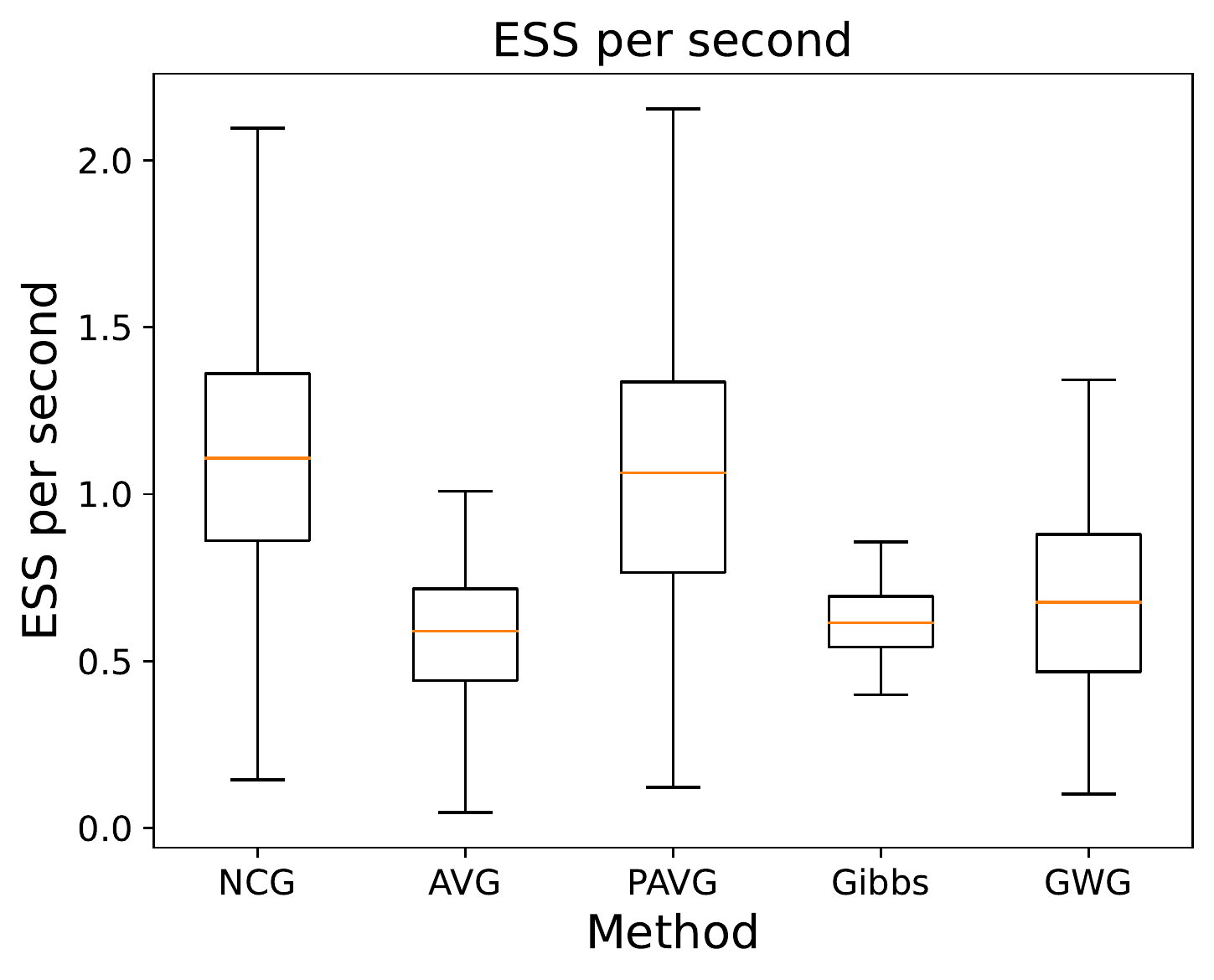}
\end{subfigure}
 \caption{100D Sparse Bayesian linear regression results. The proposed samplers (NCG, AVG \& PAVG) converge faster as measured by bivariate marginals (middle) and are competitive or better as measured by univariate marginals (left) and Effective Sample Size (ESS) (right).}
   \label{fig:sparse bayes 100d}
\end{figure}

A common use-case of MCMC is sampling posterior distributions that arise in Bayesian analysis. We consider a Bayesian treatment of sparse variable selection in linear regression models, using an experimental setup inspired by that of \citet{titsias2017hamming}. Given an $n \times d$ design matrix $X$, the response $\vy \in \mathbb{R}^n$ is modelled as
\begin{align}
    \vy = X (\s \odot \bm{\omega}) + \sigma \bm{\nu}  &&  \bm{\nu} \sim \mathcal{N}(0, \mI_n),
\end{align}
where $\bm{\omega} \in \mathbb{R}^d$ is a vector of weights and $\s \in \mathbb{R}^d$ is a binary random vector that masks out certain covariates. We place a conjugate normal-inverse-gamma prior over weights and noise-variance $(\bm{\omega}, \sigma^2)$, which can then be analytically marginalised out. Combined with a sparsity-promoting prior over $\s$, we obtain an unnormalised expression for the posterior over binary masks $p(\s \given X, \vy)$---see Appendix \ref{appendix: sbl posterior} for a complete description. This posterior tells us which covariates are `relevant' for predicting the response $\vy$, and which are `irrelevant'. Thus, it can be viewed as a kind of sparse variable selection procedure.


This posterior is a 20D distribution over binary vectors, implying $\sim 1$m possible states, which we can sum over (enabling normalisation). To make the sampling problem more challenging, whilst keeping normalisation tractable, we append additional `irrelevant' dimensions by multiplying the posterior with $\prod_{i=1}^{80} \mathcal{B}(0.001)$, where $\mathcal{B}$ is Bernoulli distribution. The fact that we can normalise (and sample) the resulting target distribution, $p(\s)$, enables us to compare it to the empirical distribution of MCMC samples $q(\s)$ using the following two metrics i) marginal estimation error: $(1/d)\sum_i^d|q_i(\si = 1) - p_i(\si = 1)|$ and ii) pairwise estimation error: $\frac{1}{d^2}\sum_{i, j}^d \sum_{k, l \in \{0, 1\}} |q_{i, j}(\si=k, \sj=l) - p_{i, j}(\si=k, \sj=l)|$. These metrics, as well as ESS, are described more fully in Appendix \ref{appendix: ordinal and sbl eval metrics}.

\subsubsection{Results}

Figure \ref{fig:sparse bayes 100d} shows the results. Under all evaluation metrics, the methods rank as follows: PAVG $\geq$ NCG $\geq$ AVG $\geq$ GWG $\geq$ Gibbs, with the pairwise-error metric indicating that these inequalities are strict. This ranking matches our findings from the previous experiment, except for the swapping of GWG and Gibbs. Most interestingly, we continue to see a clear benefit from preconditioning, even though the target log-probability is far from quadratic.

\subsection{Estimation of Ising models via persistent-contrastive divergence}
\label{sec: ising experiment}

\begin{figure}
\setlength{\LTpre}{0pt}
\LTcapwidth=\textwidth
\setlength{\tabcolsep}{5pt}
\begin{longtable}[t]{@{}l r>{\footnotesize}r r>{\footnotesize}r r>{\footnotesize}r r>{\footnotesize}r r>{\footnotesize}r}
\caption{Estimation error of an Ising Lattice matrix learned with persistent contrastive divergence. Error bars are standard deviations across 5 runs. Each run uses a different seed for all sources of randomness.}\label{table: ising table} \\

\toprule%
 \centering%
 & \multicolumn{2}{c}{{{ K=1$^*$}}}
 & \multicolumn{2}{c}{{{ K=5}}}
 & \multicolumn{2}{c}{{{ K=10}}}
 & \multicolumn{2}{c}{{{ K=15}}}
 & \multicolumn{2}{c}{{{ K=20}}} \\

\cmidrule[0.3pt](r{0.125em}){1-1}%
\cmidrule[0.3pt](lr{0.125em}){2-3}%
\cmidrule[0.3pt](lr{0.125em}){4-5}%
\cmidrule[0.3pt](lr{0.125em}){6-7}%
\cmidrule[0.3pt](lr{0.125em}){8-9}%
\cmidrule[0.3pt](lr{0.125em}){10-11}%
\endhead

NCG & 
0.837 & ($\pm.05$) & 
\highest{0.117} & ($\pm.0003$) & 
\highest{0.117} & ($\pm.0004$) & 
\highest{0.117} & ($\pm.0005$) & 
\highest{0.117} & ($\pm.0005$) \\

\myrowcolour%
AVG & 
5.535 & ($\pm.002$) & 
5.452 & ($\pm.003$) & 
5.273 & ($\pm.007$) & 
4.814 & ($\pm.029$) & 
0.120 & ($\pm.0006$) \\

PAVG \small{model-specific}\textsuperscript{\Cross} & 
\highest{0.120} & ($\pm.003$) & 
\highest{0.118} & ($\pm.0003$) & 
\highest{0.118} & ($\pm.0003$) & 
\highest{0.118} & ($\pm.0004$) & 
\highest{0.118} & ($\pm.0005$) \\

\myrowcolour%
PAVG \small{model-agnostic} & 
5.521 & ($\pm.002$) & 
0.124 & ($\pm0.0005$) & 
0.121 & ($\pm0.0004$) & 
\highest{0.119} & ($\pm0.0004$) & 
\highest{0.119} & ($\pm0.0007$) \\

GWG &
5.271 & ($\pm.0038$) & 
0.163 & ($\pm.0009$) & 
0.138 & ($\pm.0004$) & 
0.132 & ($\pm.0002$) & 
0.128 & ($\pm.0008$) \\

\myrowcolour%
Gibbs &
4.825 & ($\pm.002$) & 
0.805 & ($\pm.003$) & 
0.167 & ($\pm.0005$) & 
0.136 & ($\pm.0003$) & 
0.132 & ($\pm.0005$) \\

\bottomrule
    \multicolumn{11}{l}{ \small{$^*$ K refers to the number of MCMC steps used by \emph{AVG}; suitable multipliers are used so each method has the same budget.}} \\
    \multicolumn{11}{l}{ \small{\textsuperscript{\Cross} This corresponds precisely to the Block-Gibbs sampler introduced by \citet{martens2010parallelizable}}}

\end{longtable}
\end{figure}

Pairwise undirected graphical models are an important class of distributions used in physics, proteomics and economics \citep{mackay2003information, lapedes1999correlated, sornette2014physics}. Here, we consider the Ising model
\begin{align}
    \log p(\s) &= \rvb^T\s + \frac{1}{2} \s^T J \s - \log Z, & \s \in \{-1, 1\}^d
\end{align}
where $J$ is a binary matrix multiplied by a constant and $Z$ is the (generally intractable) normaliser.

Following \citet{grathwohl2021oops}, we define a 100-dimensional Ising model with ground-truth parameters $\rvb^*=0$ and $J^*$ is set to a cyclic lattice as depicted in Appendix \ref{appendix: ising lattice}. We then obtain `ground-truth' samples from this model by running a Gibbs sampler for one million iterations. These samples are then used as a dataset from which we re-estimate the Ising model using Persistent Contrastive Divergence (PCD) \citep{neal1992connectionist, younes1999convergence, tieleman2008training, du2019implicit}, which is an approximation to gradient-based maximum likelihood learning that requires an MCMC sampler; see pseudocode in Algorithm \ref{algo: PCD}. 

PCD has two key free-parameters: the MCMC sampler itself and the number of sampling steps $K$ per parameter update. Better samplers enable lower values of $K$ to obtain a desired level of estimation error. For the Ising model, we quantify estimation error in terms of Frobenius norm $\norm{J - J^*}_F$, where $J$ is a matrix of free parameters, and the model's bias $\rvb$ is fixed at the ground-truth value.



\noindent \textbf{PAVG for log-quadratic targets.} \\[1ex] For this experiment, we compare two types of PAVG: model-specific and model-agnostic. The latter is simply the approach we used in previous experiments as discussed in Section \ref{sec: choice of precon mat}. The model-specific version uses $J$ as the preconditioning matrix. As described at the end of \ref{sec: pavg}, this renders PAVG identical to the log-quadratic Block-Gibbs sampler of \citet{martens2010parallelizable}.

\subsubsection{Results}
Table \ref{table: ising table} shows the results. As one might expect, the model-specific method of \citet{martens2010parallelizable} performs best, achieving low estimation error even when $K=1$. The superiority of this model-specific sampler over GWG is a new finding; indeed \citet{grathwohl2021oops} perform multiple experiments with log-quadratic target distributions but do not compare to \citet{martens2010parallelizable}.

\begin{figure}
\setlength{\LTpre}{0pt}
\LTcapwidth=\linewidth
\setlength{\tabcolsep}{12pt}
\begin{longtable}[t]{@{}l 
r>{\footnotesize}r
r>{\footnotesize}r r>{\footnotesize}r r>{\footnotesize}r}

\caption{Estimation error of quadratic term when sampling from a quadratic-ConvEBM product distribution. Error bars are standard deviations across 5 runs. Each run uses a different seed for all sources of randomness.} \label{table: conv table} \\

\toprule%
 \centering%
 & \multicolumn{2}{c}{{{ K=5$^*$}}}
 & \multicolumn{2}{c}{{{ K=10}}}
 & \multicolumn{2}{c}{{{ K=15}}}
 & \multicolumn{2}{c}{{{ K=20}}} \\

\cmidrule[0.3pt](r{0.125em}){1-1}%
\cmidrule[0.3pt](lr{0.125em}){2-3}%
\cmidrule[0.3pt](lr{0.125em}){4-5}%
\cmidrule[0.3pt](lr{0.125em}){6-7}%
\cmidrule[0.3pt](lr{0.125em}){8-9}%
\endhead

NCG & 
\highest{0.128} & ($\pm.001$) & 
\highest{0.116} & ($\pm.0002$) & 
\highest{0.112} & ($\pm.0009$) & 
\highest{0.112} & ($\pm.0004$) \\

\myrowcolour%
AVG & 
3.310 & ($\pm.078$) & 
2.599 & ($\pm.051$) & 
0.496 & ($\pm.119$) & 
\highest{0.114} & ($\pm.0006$) \\

PAVG & 
3.212 & ($\pm.079$) & 
\highest{0.117} & ($\pm.0004$) & 
\highest{0.114} & ($\pm.0006$) & 
\highest{0.112} & ($\pm.0007$) \\

\myrowcolour%
GWG &
0.732 & ($\pm.013$) & 
0.156 & ($\pm.003$) & 
0.120 & ($\pm.001$) & 
\highest{0.114} & ($\pm.0005$) \\

Gibbs &
3.121 & ($\pm.146$) & 
2.270 & ($\pm.067$) & 
1.637 & ($\pm.007$) & 
1.204 & ($\pm.008$) \\

\bottomrule

\multicolumn{9}{l}{ \small{$^*$ K = number of MCMC steps for \emph{AVG}; suitable multipliers are used so each method has the same budget.}} \\

\end{longtable}
\end{figure}

Amongst the non model-specific methods, the results imply the following ranking: NCG > PAVG (model-agnostic) > GWG > Gibbs > AVG.
In contrast to previous experiments, we now see a modest but clear advantage for NCG over PAVG, and that AVG underperforms all other samplers. It's important to note that, throughout learning, the $\epsilon$ step-size parameter used by NCG and (P)AVG is held fixed. This means that the same step-size must be effective across a \emph{range} of target distributions (since the target changes every time we update the model's parameters). The results show that NCG and PAVG work robustly with a fixed step-size, whilst AVG is less robust. We suspect that all three methods would benefit from adaptively-tuning the step-size throughout learning, but leave this possibility to future work.

\subsection{Sampling deep convolutional energy-based models}

Deep energy-based models (EBMs) take the form $\log p(\s) = \exp(f_{\theta}(\s)) - \log Z$, where $f_{\theta}$ is parameterised as a deep neural network. Such models have attracted significant attention for continuous data \citep{du2019implicit, arbel2020generalized, qin2022cold} where Langevin-based samplers are the default approach. In the discrete setting, less progress has been made, with \citet{grathwohl2021oops} (GWG) being the first paper to showcase the potential of high-dimensional discrete EBMs.

A major challenge in comparing the efficacy of different samplers for deep EBMs is the lack of an easy-to-compute evaluation metric. Unlike the Ising model, parameter estimation error is not meaningful since the models are not identifiable. We propose a novel strategy for dealing with this problem consisting of the following steps: 
\\[1ex]
\textbf{i)} Given a real dataset, fit a ground-truth `quadratic-EBM' product distribution $f_{J^*, \theta^*}(\s) := \frac{1}{2} \s^T J^* \s + f_{\theta^*}(\s)$
to the data using PCD with GWG and a large number of sampling steps K (we use 50).
\\[1ex]
\textbf{ii)} Sample a `ground truth' dataset $\mathcal{D}_{J^*, \theta^*}$ from the quadratic-EBM by running GWG for many (e.g.\ 50k) iterations.
\\[1ex]
\textbf{iii)} Freeze the neural parameters $\theta^*$, and re-estimate $J$ from the dataset $\mathcal{D}_{J^*, \theta^*}$ using PCD with any choice of MCMC sampler and value of $K$. Holding $\theta^*$ fixed means $J$ is identifiable, and the estimation error, $\norm{J - J^*}_F$, is a valid performance metric.
\\[1ex]
We apply this methodology to the USPS 256-dimensional image dataset of binarised handwritten digits \citep{hull1994database} and parameterise $f_{\theta}$ as a 7-layer convolutional network as detailed in Appendix \ref{appendix: conv ebm}.




\subsubsection{Results}

Table \ref{table: conv table} shows the results. Roughly speaking, we can rank the methods as $\text{NCG > PAVG} \approx \text{GWG > AVG}  \approx \text{Gibbs}$
where the $\approx$ symbol indicates that neither method is clearly superior. This ranking is similar that obtained in the previous experiment, except that PAVG is no longer clearly superior to GWG and Gibbs is no longer clearly superior to AVG. We note that the estimation errors in Table \ref{table: conv table} correlate well with visual sample quality; see images in the appendix, Figure \ref{fig: mcmc buffers per method for USPS quadratic-EBM}.

\section{Discussion}
We have presented multiple discrete gradient-based MCMC samplers that show strong performance across a range of problem types in Bayesian inference and energy-based modelling. In particular, we obtained the NCG sampler by viewing the Metropolis-adjusted Langevin Algorithm through the lens of locally-informed proposals \citep{zanella2020informed} and the PAVG sampler through the lens of gradient-based auxiliary samplers \citet{titsias2018auxiliary}.

Depending on the task, we saw that either NCG or PAVG show the strongest performance, and both generally outperform Gibbs-with-Gradients (GWG) by a clear margin. This demonstrates the value of using proposal distributions that update multiple dimensions at once, and do so in a correlated way. However, it is important to note that these advantages do not come for free: GWG requires no tuning, whereas both NCG and PAVG have step-size parameters. In practice, we would recommend running GWG alongside the new methods when facing a new sampling problem, as it provides a strong and reliable baseline.

To our knowledge, we are the first to adapt the auxiliary variable framework of \citet{titsias2018auxiliary} to discrete state-spaces. We believe there is significant scope to build on this idea, by investigating different choices of continuous conditional distributions, and understanding how/when \emph{marginalised} auxiliary proposals can be leveraged. In the context of PAVG, a natural question is what happens if one replaces the \emph{global} preconditioning matrix with a state-specific matrix such as the Hessian, in a similar vein to Manifold MALA \citep{girolami2011riemann}.


\subsubsection*{Acknowledgments}
Benjamin Rhodes was supported in part by the EPSRC Centre for Doctoral Training in Data Science, funded by the UK Engineering and Physical Sciences Research Council (grant EP/L016427/1) and the University of Edinburgh.

\bibliography{main}

\begin{thebibliography}{40}
\providecommand{\natexlab}[1]{#1}
\providecommand{\url}[1]{\texttt{#1}}
\expandafter\ifx\csname urlstyle\endcsname\relax
  \providecommand{\doi}[1]{doi: #1}\else
  \providecommand{\doi}{doi: \begingroup \urlstyle{rm}\Url}\fi

\bibitem[Andrieu \& Thoms(2008)Andrieu and Thoms]{andrieu2008tutorial}
Christophe Andrieu and Johannes Thoms.
\newblock A tutorial on adaptive mcmc.
\newblock \emph{Statistics and computing}, 18\penalty0 (4):\penalty0 343--373,
  2008.

\bibitem[Arbel et~al.(2020)Arbel, Zhou, and Gretton]{arbel2020generalized}
Michael Arbel, Liang Zhou, and Arthur Gretton.
\newblock Generalized energy based models.
\newblock \emph{arXiv preprint arXiv:2003.05033}, 2020.

\bibitem[Dinh et~al.(2017)Dinh, Bilge, Zhang, and
  Matsen~IV]{dinh2017probabilistic}
Vu~Dinh, Arman Bilge, Cheng Zhang, and Frederick~A Matsen~IV.
\newblock Probabilistic path hamiltonian monte carlo.
\newblock In \emph{International Conference on Machine Learning}, pp.\
  1009--1018. PMLR, 2017.

\bibitem[Du \& Mordatch(2019)Du and Mordatch]{du2019implicit}
Yilun Du and Igor Mordatch.
\newblock Implicit generation and modeling with energy based models.
\newblock \emph{Advances in Neural Information Processing Systems}, 32, 2019.

\bibitem[Duane et~al.(1987)Duane, Kennedy, Pendleton, and
  Roweth]{duane1987hybrid}
Simon Duane, Anthony~D Kennedy, Brian~J Pendleton, and Duncan Roweth.
\newblock Hybrid monte carlo.
\newblock \emph{Physics letters B}, 195\penalty0 (2):\penalty0 216--222, 1987.

\bibitem[Dwivedi et~al.(2018)Dwivedi, Chen, Wainwright, and Yu]{dwivedi2018log}
Raaz Dwivedi, Yuansi Chen, Martin~J Wainwright, and Bin Yu.
\newblock Log-concave sampling: Metropolis-hastings algorithms are fast!
\newblock In \emph{Conference on learning theory}, pp.\  793--797. PMLR, 2018.

\bibitem[Geman \& Geman(1984)Geman and Geman]{geman1984stochastic}
Stuart Geman and Donald Geman.
\newblock Stochastic relaxation, gibbs distributions, and the bayesian
  restoration of images.
\newblock \emph{IEEE Transactions on pattern analysis and machine
  intelligence}, \penalty0 (6):\penalty0 721--741, 1984.

\bibitem[Girolami \& Calderhead(2011)Girolami and
  Calderhead]{girolami2011riemann}
Mark Girolami and Ben Calderhead.
\newblock Riemann manifold langevin and hamiltonian monte carlo methods.
\newblock \emph{Journal of the Royal Statistical Society: Series B (Statistical
  Methodology)}, 73\penalty0 (2):\penalty0 123--214, 2011.

\bibitem[Grathwohl et~al.(2021)Grathwohl, Swersky, Hashemi, Duvenaud, and
  Maddison]{grathwohl2021oops}
Will Grathwohl, Kevin Swersky, Milad Hashemi, David Duvenaud, and Chris
  Maddison.
\newblock Oops i took a gradient: Scalable sampling for discrete distributions.
\newblock In \emph{International Conference on Machine Learning}, pp.\
  3831--3841. PMLR, 2021.

\bibitem[Hastings(1970)]{hastings1970monte}
W~Keith Hastings.
\newblock Monte carlo sampling methods using markov chains and their
  applications.
\newblock 1970.

\bibitem[Hertz et~al.(1991)Hertz, Krogh, Palmer, and
  Horner]{hertz1991introduction}
John Hertz, Anders Krogh, Richard~G Palmer, and Heinz Horner.
\newblock Introduction to the theory of neural computation.
\newblock \emph{Physics Today}, 44\penalty0 (12):\penalty0 70, 1991.

\bibitem[Hubbard(1959)]{hubbard1959calculation}
John Hubbard.
\newblock Calculation of partition functions.
\newblock \emph{Physical Review Letters}, 3\penalty0 (2):\penalty0 77, 1959.

\bibitem[Hull(1994)]{hull1994database}
Jonathan~J. Hull.
\newblock A database for handwritten text recognition research.
\newblock \emph{IEEE Transactions on pattern analysis and machine
  intelligence}, 16\penalty0 (5):\penalty0 550--554, 1994.

\bibitem[Kingma \& Ba(2014)Kingma and Ba]{kingma2014adam}
Diederik~P Kingma and Jimmy Ba.
\newblock Adam: A method for stochastic optimization.
\newblock \emph{arXiv preprint arXiv:1412.6980}, 2014.

\bibitem[Lapedes et~al.(1999)Lapedes, Giraud, Liu, and
  Stormo]{lapedes1999correlated}
Alan~S Lapedes, Bertrand~G Giraud, LonChang Liu, and Gary~D Stormo.
\newblock Correlated mutations in models of protein sequences: phylogenetic and
  structural effects.
\newblock \emph{Lecture Notes-Monograph Series}, pp.\  236--256, 1999.

\bibitem[Levy et~al.(2018)Levy, Hoffman, and
  Sohl-Dickstein]{levy2018generalizing}
Daniel Levy, Matt~D Hoffman, and Jascha Sohl-Dickstein.
\newblock Generalizing hamiltonian monte carlo with neural networks.
\newblock In \emph{International Conference on Learning Representations}, 2018.

\bibitem[Ma et~al.(2015)Ma, Chen, and Fox]{ma2015complete}
Yi-An Ma, Tianqi Chen, and Emily Fox.
\newblock A complete recipe for stochastic gradient mcmc.
\newblock \emph{Advances in neural information processing systems}, 28, 2015.

\bibitem[MacKay et~al.(2003)MacKay, Mac~Kay, et~al.]{mackay2003information}
David~JC MacKay, David~JC Mac~Kay, et~al.
\newblock \emph{Information theory, inference and learning algorithms}.
\newblock Cambridge university press, 2003.

\bibitem[Martens \& Sutskever(2010)Martens and
  Sutskever]{martens2010parallelizable}
James Martens and Ilya Sutskever.
\newblock Parallelizable sampling of markov random fields.
\newblock In \emph{Proceedings of the Thirteenth International Conference on
  Artificial Intelligence and Statistics}, pp.\  517--524. JMLR Workshop and
  Conference Proceedings, 2010.

\bibitem[Metropolis et~al.(1953)Metropolis, Rosenbluth, Rosenbluth, Teller, and
  Teller]{metropolis1953equation}
Nicholas Metropolis, Arianna~W Rosenbluth, Marshall~N Rosenbluth, Augusta~H
  Teller, and Edward Teller.
\newblock Equation of state calculations by fast computing machines.
\newblock \emph{The journal of chemical physics}, 21\penalty0 (6):\penalty0
  1087--1092, 1953.

\bibitem[Neal(1992)]{neal1992connectionist}
Radford~M Neal.
\newblock Connectionist learning of belief networks.
\newblock \emph{Artificial intelligence}, 56\penalty0 (1):\penalty0 71--113,
  1992.

\bibitem[Neal et~al.(2011)]{neal2011mcmc}
Radford~M Neal et~al.
\newblock Mcmc using hamiltonian dynamics.
\newblock \emph{Handbook of markov chain monte carlo}, 2\penalty0
  (11):\penalty0 2, 2011.

\bibitem[Nishimura et~al.(2017)Nishimura, Dunson, and
  Lu]{nishimura2017discontinuous}
Akihiko Nishimura, David Dunson, and Jianfeng Lu.
\newblock Discontinuous hamiltonian monte carlo for sampling discrete
  parameters.
\newblock \emph{arXiv preprint arXiv:1705.08510}, 2, 2017.

\bibitem[Pakman \& Paninski(2013)Pakman and Paninski]{pakman2013auxiliary}
Ari Pakman and Liam Paninski.
\newblock Auxiliary-variable exact hamiltonian monte carlo samplers for binary
  distributions.
\newblock \emph{Advances in neural information processing systems}, 26, 2013.

\bibitem[Qin et~al.(2022)Qin, Welleck, Khashabi, and Choi]{qin2022cold}
Lianhui Qin, Sean Welleck, Daniel Khashabi, and Yejin Choi.
\newblock Cold decoding: Energy-based constrained text generation with langevin
  dynamics.
\newblock \emph{arXiv preprint arXiv:2202.11705}, 2022.

\bibitem[Ramachandran et~al.(2017)Ramachandran, Zoph, and
  Le]{ramachandran2017searching}
Prajit Ramachandran, Barret Zoph, and Quoc~V Le.
\newblock Searching for activation functions.
\newblock \emph{arXiv preprint arXiv:1710.05941}, 2017.

\bibitem[Robert et~al.(1999)Robert, Casella, and Casella]{robert1999monte}
Christian~P Robert, George Casella, and George Casella.
\newblock \emph{Monte Carlo statistical methods}, volume~2.
\newblock Springer, 1999.

\bibitem[Roberts \& Rosenthal(1998)Roberts and Rosenthal]{roberts1998optimal}
Gareth~O Roberts and Jeffrey~S Rosenthal.
\newblock Optimal scaling of discrete approximations to langevin diffusions.
\newblock \emph{Journal of the Royal Statistical Society: Series B (Statistical
  Methodology)}, 60\penalty0 (1):\penalty0 255--268, 1998.

\bibitem[Roberts \& Tweedie(1996)Roberts and Tweedie]{roberts1996exponential}
Gareth~O Roberts and Richard~L Tweedie.
\newblock Exponential convergence of langevin distributions and their discrete
  approximations.
\newblock \emph{Bernoulli}, pp.\  341--363, 1996.

\bibitem[Rosenthal et~al.(2011)]{rosenthal2011optimal}
Jeffrey~S Rosenthal et~al.
\newblock Optimal proposal distributions and adaptive mcmc.
\newblock \emph{Handbook of Markov Chain Monte Carlo}, 4\penalty0 (10.1201),
  2011.

\bibitem[Sornette(2014)]{sornette2014physics}
Didier Sornette.
\newblock Physics and financial economics (1776--2014): puzzles, ising and
  agent-based models.
\newblock \emph{Reports on progress in physics}, 77\penalty0 (6):\penalty0
  062001, 2014.

\bibitem[Tieleman(2008)]{tieleman2008training}
Tijmen Tieleman.
\newblock Training restricted boltzmann machines using approximations to the
  likelihood gradient.
\newblock In \emph{Proceedings of the 25th international conference on Machine
  learning}, pp.\  1064--1071, 2008.

\bibitem[Titsias \& Papaspiliopoulos(2018)Titsias and
  Papaspiliopoulos]{titsias2018auxiliary}
Michalis~K Titsias and Omiros Papaspiliopoulos.
\newblock Auxiliary gradient-based sampling algorithms.
\newblock \emph{Journal of the Royal Statistical Society: Series B (Statistical
  Methodology)}, 80\penalty0 (4):\penalty0 749--767, 2018.

\bibitem[Titsias \& Yau(2017)Titsias and Yau]{titsias2017hamming}
Michalis~K Titsias and Christopher Yau.
\newblock The hamming ball sampler.
\newblock \emph{Journal of the American Statistical Association}, 112\penalty0
  (520):\penalty0 1598--1611, 2017.

\bibitem[Vehtari et~al.(2021)Vehtari, Gelman, Simpson, Carpenter, and
  B{\"u}rkner]{vehtari2021rank}
Aki Vehtari, Andrew Gelman, Daniel Simpson, Bob Carpenter, and Paul-Christian
  B{\"u}rkner.
\newblock Rank-normalization, folding, and localization: an improved r for
  assessing convergence of mcmc (with discussion).
\newblock \emph{Bayesian analysis}, 16\penalty0 (2):\penalty0 667--718, 2021.

\bibitem[Younes(1999)]{younes1999convergence}
Laurent Younes.
\newblock On the convergence of markovian stochastic algorithms with rapidly
  decreasing ergodicity rates.
\newblock \emph{Stochastics: An International Journal of Probability and
  Stochastic Processes}, 65\penalty0 (3-4):\penalty0 177--228, 1999.

\bibitem[Zanella(2020)]{zanella2020informed}
Giacomo Zanella.
\newblock Informed proposals for local mcmc in discrete spaces.
\newblock \emph{Journal of the American Statistical Association}, 115\penalty0
  (530):\penalty0 852--865, 2020.

\bibitem[Zellner(1986)]{zellner1986assessing}
Arnold Zellner.
\newblock On assessing prior distributions and bayesian regression analysis
  with g-prior distributions.
\newblock \emph{Bayesian inference and decision techniques}, 1986.

\bibitem[Zhang et~al.(2022)Zhang, Liu, and Liu]{zhang2022langevin}
Ruqi Zhang, Xingchao Liu, and Qiang Liu.
\newblock A langevin-like sampler for discrete distributions.
\newblock In \emph{International Conference on Machine Learning}, pp.\
  26375--26396. PMLR, 2022.

\bibitem[Zhang et~al.(2012)Zhang, Ghahramani, Storkey, and
  Sutton]{zhang2012continuous}
Yichuan Zhang, Zoubin Ghahramani, Amos~J Storkey, and Charles Sutton.
\newblock Continuous relaxations for discrete hamiltonian monte carlo.
\newblock \emph{Advances in Neural Information Processing Systems}, 25, 2012.

\end{thebibliography}
\bibliographystyle{tmlr}

\newpage
\appendix

\section{One-hot categorical random variables}
\label{appendix: categorical vars}

A categorical random variable with $k$ possible values can be represented as a one-hot vector belonging to the set
\vspace{-0.5em}
\begin{align}
    & \mathcal{S} = \{\s : \si \in \{0, 1\}, \sum_{i=1}^k s_i = 1 \} \subset \{0, 1\}^k & \text{where} \ \ |\mathcal{S}| = k
\end{align}
A $d$-length vector of $k$-valued categorical random variables thus belongs to the set $\mathcal{S}^d \subset \mathbb{R}^{dk}$. 

The methods introduced in this paper (NCG, AVG and PAVG) all use proposal distributions that, before normalisation, have the form $\exp( \va^T\s + \vb^T (\s \odot \s)$ for some choice of $\va$ and $\vb$. Such formulas remain valid in the categorical case. However, the \emph{normalised} versions of these distributions have a slightly different form. Specifically, to normalise this distribution, we note that there are $d$-independent groups of dimensions, and that within each group, there are only $k$ possible settings, yielding
\vspace{-0.5em}
\begin{align}
& \prod_{j=1}^{d} \sigma( \va_j^T\s_j + \vb_j^T (\s_j \odot \s_j)) &  \sigma(\ervx) = \frac{\exp(\ervx)}{\sum_{\mathcal{S}} \exp(\ervx)}, \label{eq: appendix categorical formula}
\end{align}
where $\s_j := \s_{(j-1)k:jk}$, $\va_j := \va_{(j-1)k:jk}$ and $\vb_j := \vb_{(j-1)k:jk}$.

\section{Sampling algorithms}
\label{appendix: sampling pseudocode}

\begin{algorithm}
{\fontsize{11}{14}\selectfont
\caption{NCG step}\label{algo: NCG}
\begin{algorithmic}
\Require Step-size $\epsilon$. Unnormalised log prob function $f(\cdot)$. Triple $(\st, f(\st), \nabla f(\st) )$.
\State Sample $\stt \sim q_{\epsilon}(\rvs \given \st)$ as in Eq. \ref{eq: ncg local-proposal factorised} (binary/ordinal) or the categorical equivalent implied by Eq. \ref{eq: appendix categorical formula}
\State Compute $f(\stt)$ \& $\nabla f(\stt)$
\State Accept $\stt$ with probability \vspace{-2.7em} \begin{align}
    \min \Big( 1, \exp(f(\stt) -  f(\st) )\frac{q_{\epsilon}(\st \given \stt)}{q_{\epsilon}(\stt \given \st)} \Big)
\end{align}
\end{algorithmic}
}
\end{algorithm}

\begin{algorithm}
{\fontsize{11}{14}\selectfont
\caption{AVG step}\label{algo: AVG}
\begin{algorithmic}
\Require Step-size $\epsilon$. Unnormalised log prob function $f(\cdot)$. Triple $(\st, f(\st), \nabla f(\st) )$.
\State Sample $\zt \sim \mathcal{N}(\z; \  \sqrt{2/\epsilon} \ \st)$
\State Sample $\stt \sim q_{\epsilon}(\rvs \given \zt, \st)$ as in Eq. \ref{eq: avg univariate factor} (binary/ordinal) or the categorical equivalent implied by Eq. \ref{eq: appendix categorical formula}
\State Compute $f(\stt)$ \& $\nabla f(\stt)$
\State Accept $\stt$ with probability \vspace{-1em} \begin{align}
    \min \Big( 1, \exp(f(\stt) -  f(\st) ) \frac{\mathcal{N}(\z; \  \sqrt{2/\epsilon} \ \stt)}{\mathcal{N}(\z; \  \sqrt{2/\epsilon} \ \st)} \frac{q_{\epsilon}(\st \given \zt, \stt)}{q_{\epsilon}(\stt \given \zt, \st)} \Big)
\end{align}
\end{algorithmic}
}
\end{algorithm}

\begin{algorithm}
{\fontsize{11}{14}\selectfont
\caption{PAVG step}\label{algo: PAVG}
\begin{algorithmic}
\Require Step-size $\epsilon$. Preconditioner $\Sigma$. Unnormalised log prob function $f(\cdot)$. Triple $(\st, f(\st), \nabla f(\st) )$.
\State Sample $\zt \sim \mathcal{N}(\z; \ \Sigeps^{1/2}\st, \mI)$ for $\Sigeps$ defined in Eq. \ref{eq: appendix sigmaeps with neg eigval}.
\State Sample $\stt \sim q_{\epsilon}(\rvs \given \zt, \st)$ as in Eq. \ref{eq: appendix pavg neg eigval proposal} (binary/ordinal) or the categorical equivalent implied by Eq. \ref{eq: appendix categorical formula}
\State Compute $f(\stt)$ \& $\nabla f(\stt)$
\State Accept $\stt$ with probability \vspace{-1em} \begin{align}
    \min \Big( 1, \exp(f(\stt) -  f(\st) ) \frac{\mathcal{N}(\z; \ \Sigeps^{1/2}\stt, \mI)}{\mathcal{N}(\z; \ \Sigeps^{1/2}\st, \mI)} \frac{q_{\epsilon}(\st \given \zt, \stt)}{q_{\epsilon}(\stt \given \zt, \st)} \Big)
\end{align}
\end{algorithmic}
}
\end{algorithm}


\section{Preconditioned MALA as an auxiliary variable scheme}
\label{appendix: aux var precon mala}
Here, we explain how to frame preconditioned MALA (\eqref{eq: pmala}) as an auxiliary variable sampler, and why discretising it is problematic.

For a continuous state $\s \in \mathbb{R}^d$, consider the unnormalised target density $ \pi(\s, \z) = \exp(f(\s)) \mathcal{N}(\z; \Sigma^{-1/2}\s, \mI)$. We approximate $\pi(\s \given \zt) \propto \pi(\s, \zt)$ by the proposal distribution
\begin{align}
    q_{\epsilon}(\s \given \zt, \st)  &\propto \exp(f(\st) + \nabla f(\st)^T (\s - \st)) \mathcal{N}(\zt; \Sigma^{-1/2}\s, \mI) \\
    & \propto \exp \Big( -\frac{1}{2} ( \s - \bm{\mu})^T \Sigma^{-1} (\s - \bm{\mu}) \Big) \hspace{5em} \text{where} \ \ \bm{\mu} := \Sigma^{1/2} \zt + \Sigma \nabla f(\st) \label{eq: appendix: pmala unnormalised auxiliary var proposal} \\
    &= \mathcal{N}(\s; \ \Sigma^{1/2} \zt + \Sigma \nabla f(\st), \Sigma)
\end{align}
If we now marginalise out the latents, we obtain
\begin{align}
    \int \mathcal{N}(\zt; \Sigma^{-1}\st, \mI) \mathcal{N}(\s; \ \Sigma^{1/2} \zt + \Sigma \nabla f(\st), \Sigma) d \zt = \mathcal{N}(\s; \ \st + \Sigma \nabla f(\st), 2 \Sigma)
\end{align}
which is equal to the PMALA proposal in \eqref{eq: pmala} after redefining $\Sigma$ as $\frac{\epsilon}{2} \Sigma$. Unfortunately, if we now assume $\s \in \mathcal{S}^d \subset \mathbb{R}^d$, then \eqref{eq: appendix: pmala unnormalised auxiliary var proposal} is no longer proportional to a Gaussian distribution, but rather a discrete pairwise Markov random field that is generally intractable to normalise and sample.

\section{Choice of preconditioning matrix for PAVG}
\label{appendix: choice of precon mat}

We use matrices of the form $\gamma \Sigma$, where $\gamma$ is an adaptively learned scaling parameter. In general settings, when we have no domain-knowledge to help us select $\Sigma$, we choose it to be an empirical covariance/precision matrix (see next section for how to choose) computed from a set of initial samples obtained during the burn-in phase. The algorithm for this general case is presented in \ref{algo: preconditioning matrix}.

However, when sampling from energy-based models (EBMs), we usually have access to a real-world dataset that the EBM is modelling. In this case, we can compute the empirical covariance/precision of this dataset, and use that as our $\Sigma$. This simplifies \ref{algo: preconditioning matrix}: we drop lines \ref{algo: preconmat sigma start line}-\ref{algo: preconmat sigma end line}, and no longer need a zero-initialisation of $\Sigma$ in line \ref{algo: preconmat init sigma line}.

\subsection{Covariance or precision? Automatically picking $\Sigma$ from a list of options.}
\label{appendix: fitpreconmatrix}
If we were sampling continuous variables drawn from a Gaussian distribution, then the correct choice of $\Sigma$, based on the approximation we made in \eqref{eq: global quad approx}, would be the (negative) precision matrix. Not too surprisingly, this choice works best in our ordinal experiments in Section \ref{sec: ordinal experiments}. However, in all of our binary experiments, the \emph{covariance} matrix, not precision, works better (often substantially so).

The choice between covariance and precision (or, more generally, a list of candidate matrices) is a hyperparameter. To avoid manual tuning, we propose a simple heuristic to automatically select the `best' matrix from a list of options. We define `best' as the matrix that (after rescaling) minimises the approximation error in  \eqref{eq: global quad approx} across pairs of adjacent states accumulated during a burn-in phase. Specifically, we do the following
\begin{enumerate}
    \item During the burn-in period, collect $(\st, f(\st), \nabla f(\st))$ into a `dataset' $\mathcal{D}$.
    \item Assemble a list of candidates $[\Sigma_1, \Sigma_2 \ldots \Sigma_m]$. In our experiments, we consider only two candidates: the empirical covariance \& precision matrix of collected samples $\{ \st \}$.
    \item For each $\Sigma$ in our list, solve the least-squares linear regression problem
    \begin{align}
        & \argmin_{\gamma_0} \sum_{t} \norm{y_t - \gamma_0 x_t}^2 \\
         \text{where} \hspace{2em} y_t := f(\stt) - f(\st) - & \nabla f(\st)^T  (\stt - \st),
        \hspace{2em} x_t := (1/2) (\stt - \st)^T \Sigma (\stt - \st)
    \end{align}
    \item Return the re-scaled matrix $\gamma_0 \Sigma$ which obtained the lowest least-squared loss.
\end{enumerate}
Across many of our experiments, the returned matrix $\gamma_0 \Sigma$ performed very well without any additional alterations. However, to maximise the performance of PAVG, we found it beneficial to only use the scaling factor $\gamma_0$ as an `initialisation' and continue to update it adaptively using \texttt{AdaptGamma} as shown in Algorithms \ref{algo: preconditioning matrix} and \ref{algo: adapt gamma}.

\begin{algorithm}
{\fontsize{11}{16}\selectfont
\caption{Adaptive learning of preconditioning matrix. ({\footnotesize Default values in brackets are used across \emph{all} experiments)}}\label{algo: preconditioning matrix}
\begin{algorithmic}[1]
\Require Integers $N_{\text{iters}}$, $N_{\text{chains}}$ {\small (100)}, $N_{\Sigma}$ {\small (1000)} and $N_{\text{adapt}}$ {\small (100)}.
\Require Initial adaptation rate $\delta$ {\small (0.25)} and decay factor $\rho$ {\small (0.99)}.
\State Initialise chains $S_1 = [ \s^1, \ldots, \s^{N_{\text{chains}}} ]$ and history $\mathcal{D} = [S_1]$
\State Initialise $\Sigma = 0$, $\gamma = \gamma_{\text{old}} = 1.0$ \label{algo: preconmat init sigma line}
\For{$t \in \{1, \ldots N_{\text{iters}}\}$}
\State Compute $S_{t+1}$ from $S_t$ using one step of PAVG (Algorithm \ref{algo: PAVG}) with preconditioner $\gamma \Sigma$
\If{$t < N_{\Sigma}$}  \label{algo: preconmat sigma start line}
\State Append $S_{t+1}$ to history $\mathcal{D}$
\ElsIf{$t = N_{\Sigma}$}
\State Define new $\Sigma$ as described in Section (\ref{appendix: fitpreconmatrix}) using history $\mathcal{D}$ \label{algo: preconmat sigma end line}
\ElsIf{$t \mod N_{\text{adapt}} = 0$}
\State $\gamma$, $\gamma_{\text{old}}$ = \texttt{AdaptGamma}($\gamma$, $\gamma_{\text{old}}$, $\delta$, $N_{\text{adapt}}$, $\mathcal{D}$) \Comment{Update scaling factor}
\State $\delta \gets \rho \delta$  \Comment{exponential decay of step-size}
\EndIf
\EndFor

\end{algorithmic}
}
\end{algorithm}

\begin{algorithm}
{\fontsize{11}{16}\selectfont
\caption{Adapt $\gamma$ to maximise `jump' distance $\norm{\st - \s_{t-1}}_1$}\label{algo: adapt gamma}
\begin{algorithmic}
\Function{AdaptGamma}{$\gamma$, $\gamma_{\text{old}}$, $\delta$, $N_{\text{adapt}}$, $\mathcal{D}$}
\State $a_{\text{new}} \gets$  Average value of $\norm{\st - \s_{t-1}}_1$ computed over all chains in $\mathcal{D}[-N_{\text{adapt}}:]$
\State $a_{\text{old}} \gets$  Average value of $\norm{\st - \s_{t-1}}_1$ computed over all chains in $\mathcal{D}[-2N_{\text{adapt}}:-N_{\text{adapt}}]$
\State \texttt{Increased} $\gets \gamma \geq \gamma_{\text{old}}$
\State \texttt{Improved} $\gets a_{\text{new}} \geq a_{\text{old}}$
\If{(\texttt{Increased} \Aand \texttt{Improved} ) \Or (\Not \texttt{Increased} \Aand \Not\texttt{Improved})}
\State $\hat{\delta} \gets \delta$ \Comment{Positive adjustment}
\Else
\State $\hat{\delta} \gets -\delta$ \Comment{Negative adjustment}
\EndIf

\State $\gamma_{\text{old}} \gets \gamma$
\If{$|\gamma| \geq 1$
\State $\gamma \gets \gamma * (1 + \hat{\delta})$ } \Comment{Multiplicative adjustment}
\Else
\State $\gamma \gets \gamma + \hat{\delta}$ \Comment{Additive adjustment (allows $\gamma$ to change sign)}
\EndIf

\State \Return $\gamma, \gamma_{\text{old}}$

\EndFunction
\end{algorithmic}
}
\end{algorithm}

\subsection{Preconditioners with negative eigenvalues}
When deriving PAVG in section \ref{sec: pavg}, we defined a conditional Gaussian distribution of the form
\begin{align}
\label{eq: appendix pavg conditional gaussian}
&\mathcal{N}(\z; \ \Sigeps^{1/2}\st, \mI), & \Sigeps := \Sigma + (2/\epsilon) \mI.
\end{align}
We note that there are multiple types of matrix square-root, and any of them is valid here. However, such square-roots only exist if $\Sigeps$ is semi-positive definite (i.e.\ all eigenvalues are non-negative). This is only guaranteed to be the case (for any value of $\epsilon$) if $\Sigma$ is also semi-positive. This is because the eigenvalues of $\Sigeps$ are of the form $\lambda_i + (2/\epsilon)$, where $\lambda_i$ is an eigenvalue of $\Sigma$.

We can specify an alternative definition of $\Sigeps$ that ensures semi-positive definiteness
\begin{align}
\label{eq: appendix sigmaeps with neg eigval}
    \Sigeps := \Sigma + \underbrace{\big[ \max(0, -\lambda_{\text{min}}) + (2/\epsilon) \big]}_{\scalebox{1.2}{\hspace{2em} $:= d_{\epsilon}$}}\mI
\end{align}
where $\lambda_{\text{min}}$ is the most negative eigenvalue of $\Sigma$. This new definition of $\Sigeps$ means that PAVG uses a different conditional Gaussian, and the MH-proposal distribution in \eqref{eq: pavg auxiliary proposal 3} now becomes
\begin{align}
\label{eq: appendix pavg neg eigval proposal} 
    q_{\epsilon}(\s \given \zt, \st) = \prod_{i=1}^n \sigma\Big( \Big[ \nabla f(\st)_i -  (\Sigma \st)_i + (\Sigeps^{1/2} \zt)_i   \Big] \si - \frac{d_{\epsilon}}{2}  \si^2 \Big), \hspace{10mm} \text{where} \hspace{3mm} \sigma(\ervx) = \frac{\exp(\ervx)}{\sum_{\ervx \in \mathcal{S}} \exp(\ervx)},
\end{align}
What is the consequence of this new definition of $\Sigeps$? Previously, we diagonally perturbed $\Sigma$ by some value in $(0, \infty)$, where larger step-sizes $\epsilon$ meant smaller perturbations. Now, we diagonally perturb $\Sigma$ by some value in $(d_{\infty}, \infty)$, where $d_{\infty} := \max(0, -\lambda_{\text{min}}) \geq 0$. Thus compared to the old regime, the new regime enforces a kind of maximum step-size (a.k.a.\ minimum perturbation).

\section{Baseline ordinal samplers}
\label{appendix: baselines}
\begin{figure}[!b]
\centering
\includegraphics[width=.95\linewidth]{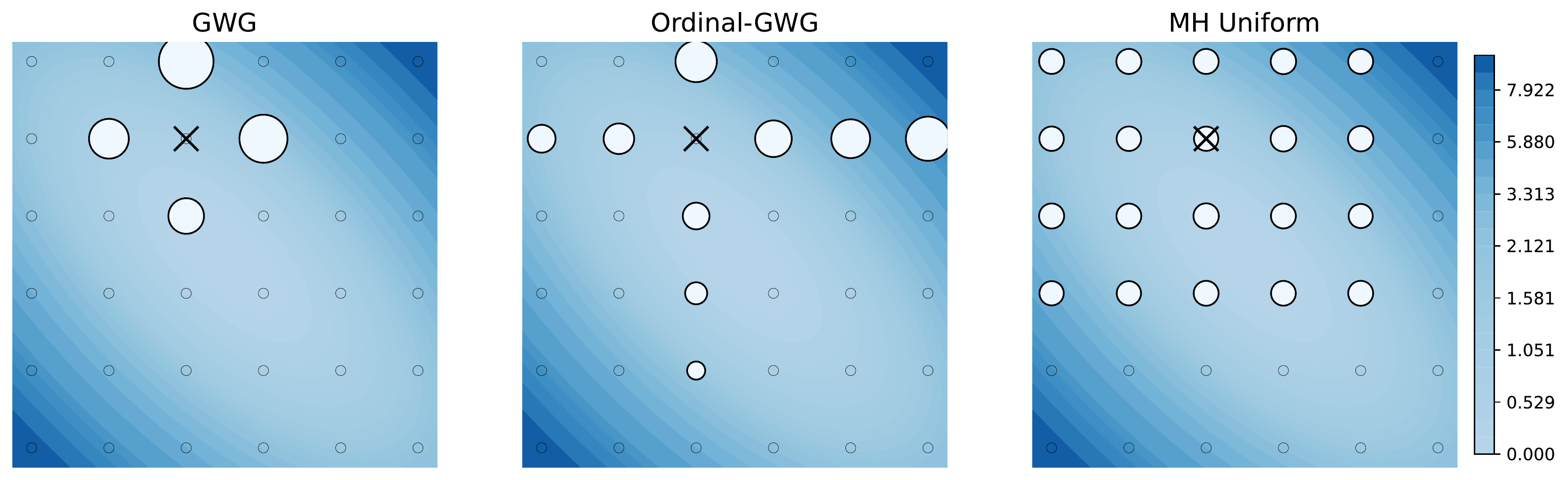}
 \caption{Illustration of the different proposal distributions used by various baseline MH samplers. The darker blue contours represent the target distribution, the light-blue dots represent the proposal distribution around the current state (black X). \textbf{Left}: GWG \textbf{Middle:} Ordinal-GWG with radius 3 \textbf{Right:} Uniform MH with radius 2.}
\label{fig: illustration of ordinal baselines}
\end{figure}
Our ordinal experiments use two additional baselines: Ordinal-GWG and MH-uniform. These are both Metropolis-Hastings samplers, with proposal distributions visualised in Figure \ref{fig: illustration of ordinal baselines}. As can be seen from the figure, Ordinal-GWG, like GWG, can only update \emph{one} dimension at a time. However, it can jump multiple states along any given dimension, with the maximum number of moves controlled by a radius parameter $r$. Thus, the support set of this proposal distribution is (at most) of size $(2r-1)d$, where $d$ is the dimensionality. The Ordinal-GWG proposal has the same functional form as GWG in \eqref{eq: gwg local-proposal}, except that the summation is no longer over a Hamming ball of radius 1, but instead over the aforementioned support set.  

MH-uniform (final column of Figure \ref{fig: illustration of ordinal baselines}) also has a radius parameter $r$. This radius controls the size of the hypercube over which the proposal is uniform.

\subsection{Tuning step-size parameters}
\label{appendix: step-size}
NCG, AVG \& PAVG have step-size parameters $\epsilon$, whilst GWG-ordinal \& MH-uniform have step-size-like parameters (a.k.a.\ radii) $r$. Our tuning procedure involves running each sampler for a short amount of time (e.g.\ 100-1000 iterations) with different step-sizes, and selecting the step-size that maximises the average L1-distance $\norm{\stt - \st}_1$ between successive states (averaged over all time-steps and parallel chains). For NCG, AVG \& PAVG we first first identify the best order-of-magnitude, and then search each decile within that particular order of magnitude e.g.\ $\{ 0.1, 0.2 \ldots, 0.9 \}$. For Ordinal-GWG and MH-uniform, we first search  $ r \in \{1, 5, 10, 15, 20\}$ and then search within the best interval e.g.\ $ r \in \{15, 16, 17, 18, 19 \}$ (note: the maximum possible value of the radius parameter is $50$ in our experiments, since this is the size of the state-space).

\begin{table}[h!]
\centering
\renewcommand{\arraystretch}{1.2}
\begin{tabular}{@{}c c c c c c@{}} 
 \toprule
 Experiment & NCG & AVG & PAVG & Ordinal-GWG & MH-Uniform \\
 \midrule
 Ordinal (poly2) & 0.05 & 0.02 & 1000.0 & 16 & 2 \\
 Ordinal (poly4) & 0.05 & 0.02 & 0.06 & 8 & 1 \\
 Bayesian regression & 0.03 & 1000.0 & 1000.0 & - & - \\
 Ising lattice & 0.5 & 0.2 & 0.2 & - & - \\
 ConvEBM & 0.2 & 0.1 & 0.1 & - & - \\
 \bottomrule
\end{tabular}
\caption{Step-sizes used across different experiments.}
\label{table:stepsizes}
\end{table}

\section{Evaluation metrics for Ordinal \& Bayesian regression experiments}
\label{appendix: ordinal and sbl eval metrics}
The experiments in \ref{sec: ordinal experiments} and \ref{sec: sparse bayes experiment} use the same methodology for evaluating performance. In both cases we:
\begin{itemize}
    \item Run 100 parallel chains for 10 minutes with a burn-in period of 1 minute.
    \item After the burn-in period, we begin saving the history of each chain.
    \item Every minute after the burn-in, we use the chain histories to compute estimation errors (Table \ref{table:estimation errors}) for each chain \emph{separately}. We then compute a mean and standard error \emph{across} our 100 parallel chains, and use these to construct Figures \ref{fig: ordinal results} and \ref{fig:sparse bayes 100d}.
    \item At the end of the run, we estimate the Effective Sample Size (ESS) of each chain. Following \citet{zanella2020informed} and \citet{grathwohl2021oops}, we map every state $\s$ in our chains to a test statistic, namely $\norm{\s - \s_{\text{random}}}_1$, where $\s_{\text{random}}$ is a randomly selected point in the state-space. The resulting test statistics can be stored in a matrix $S$ of shape \texttt{num\_iters} $\times 100$, and the ESS-per-chain is then estimated with \texttt{tfp.mcmc.effective\_sample\_size(S, filter\_beyond\_positive\_pairs=True)}. Finally, to construct the ESS plots in Figures \ref{fig: ordinal results} and \ref{fig:sparse bayes 100d}, we compute box-plot statistics (median \& quartiles) across the 100 ESS estimates.
\end{itemize}

\begin{table}[h!]
\centering
\renewcommand{\arraystretch}{1.3}
\begin{tabular}{@{}c c c @{}} 
 \toprule
 Experiment & marginal error & covariance/pairwise error \\
 \midrule
 Ordinal & $(1/d) \sum_{i=1}^d \infdiv{KL}{q_i}{p_i}$ & $\norm{\Sigma_q - \Sigma_p}_{F}$ \\
 Bayesian regression & $(1/d)\sum_i^d|q_i(\si = 1) - p_i(\si = 1)|$ & $\frac{1}{d^2}\sum_{i, j}^d \sum_{k, l \in \{0, 1\}} |q_{i, j}(\si=k, \sj=l) - p_{i, j}(\si=k, \sj=l)|.$ \\
 \bottomrule
\end{tabular}
\caption{$q_i$ refers to the i\textsuperscript{th} univariate marginal distribution of the empirical samples accumulated across a single chain. Similarly, $q_{i, j}$ refers to a bivariate marginal of such empirical samples. $p_i$ and $p_{i, j}$ refer to marginals of the target distribution, and are computed \emph{exactly} (i.e.\ they are not empirical distributions). $\Sigma_q$ and $\Sigma_p$ are both empirical covariance matrices; the former is computed using samples accumulated across a single chain, whilst the latter is computed using $100,000$ samples drawn from the target distribution.}
\label{table:estimation errors}
\end{table}

 
\section{Posterior distribution for sparse Bayesian linear regression}
\label{appendix: sbl posterior}

We define a Bayesian regression model using a similar procedure to \citet{titsias2017hamming}. The regression takes the form
\begin{align}
\label{eq: appendix bayes regression formula}
    & \vy = X (\s \odot \bm{\omega}) + \sigma \bm{\nu} & \vy, \bm{\nu} \in \mathbb{R}^{N} \ \ \s, \bm{\omega} \in \mathbb{R}^{D}
\end{align}
Where the quantities in this equation are random variables described by the following generative process
\begin{enumerate}
    \item Place a sparsity-promoting prior on $\s$ 
    \begin{align}
    \vspace{-1em}
    p(\s) = \Gamma\big(\sum_{i=1}^{D} \ervs_i + \alpha_{\pi} \big) \Gamma\big(D - \sum_{i=1}^D \ervs_i + \beta_{\pi} \big),    
    \end{align}
    where $(\alpha_{\pi}, \beta_{\pi})$ are hyperparameters with default values $(0.001, 10.0)$. This prior can itself be viewed as the marginal of a Bayesian model specified by $\s \given \pi \sim \prod_{i=1}^D \mathit{Bernoulli}(\ervs; \pi)$ and $\pi \sim \mathit{Beta}(\pi; \alpha_{\pi}, \beta_{\pi})$.
    \item Let $\bm{\nu} \sim \mathcal{N}(0, \mI_{N})$.
    \item Define $X_{\s} = X \text{diag}(\s)$ and note that $X (\s \odot \bm{\omega}) = X_{\s}\omega$.
    \item Place a conjugate normal-inverse-gamma prior over weights and noise-variance $(\bm{\omega}, \sigma^2)$. Namely,
    \begin{align}
        p(\bm{\omega}, \sigma^2 \given \s, X) = \mathcal{N}(\bm{\omega} \given 0, g\sigma^2 (X_{\s}^T X_{\s} + \lambda \mI_D)^{-1}) \ \textit{InvGamma}(\sigma^2 \given \alpha_{\sigma}, \beta_{\sigma})
    \end{align}
    We note that this choice of normal distribution is a kind of perturbed g-prior \citep{zellner1986assessing}, with hyperparameters $(g, \lambda$) with default values $(20, 0.001)$. The inverse-gamma hyperparameters $(\alpha_{\sigma}, \beta_{\sigma})$ have default values $(0.1, 0.1)$.
    \item As implied by the regression formula in \ref{eq: appendix bayes regression formula}, our model of $\vy$ follows
    \begin{align}
        p(\vy \given \s, X, \bm{\omega}, \sigma^2) = \mathcal{N}(\vy \given X_{\s} \bm{\omega}, \sigma^2 \mI_N)
    \end{align}
    \item Putting the previous steps together, we arrive at a joint distribution of the form
    \begin{align}
        p(\s, \sigma^2, \bm{\omega}, \vy \given X) = p(\s)p(\bm{\omega}, \sigma^2 \given \s, X) p(\vy \given \s, \bm{\omega}, \sigma^2, X)
    \end{align}
    The variables $(\sigma^2, \bm{\omega})$ can be analytically integrated out, resulting in the \textbf{posterior distribution}
    \begin{align}
        p(\s \given \vy, X) \propto p(\s) \ \frac{|X_{\s}^TX_{\s} + \lambda \mI_D |}{|(1+g) X_{\s}^TX_{\s} + \lambda \mI_D |}  \Big( 2 \beta_{\sigma} + \vy^T\vy - g\vy^T X_{\s} \big[ (1+g) X_{\s}^TX_{\s} + \lambda \mI_D\big] X_{\s}^T \vy \Big)^{-\frac{2\alpha_{\sigma} + N}{2}}
    \end{align}
\end{enumerate}
Finally, we use the following steps to construct `observed data' $(X, \vy)$ that we then plug into our posterior
\begin{itemize}
\item Let $\rx_1, \ldots \rx_5$ be 5 i.i.d random variables drawn from a uniform distribution over the discrete set $\{0, 1, 2\}$.
    \item Define the observed response $\evy_{\text{obs}} = \sum_{i=1}^5 \evx_i$.
    \item `Duplicate' the $5$ covariates 3 times i.e.\ $\rx_j := \rx_{j\Mod{5}+1}$ for $j \in \{6, 7 \ldots, 20\}$.
    \item Repeat the above steps $N=20$ times to obtain a `design matrix' $X \in \mathbb{R}^{20 \times 20}$ and response $\vy \in \mathbb{R}^{20}$.
\end{itemize}
 The duplication of covariates induces multi-modality in the posterior $p(\s \given X, \vy_{\text{obs}})$, since masking out $\evx_1$ and leaving its copy $\evx_6$ unmasked is equivalent to masking $\evx_6$ and leaving $\evx_1$ unmasked.
 \newpage
\section{Persistent contrastive divergence (PCD)}
\label{appendix: pcd}
\begin{algorithm}[!t]
{\fontsize{11}{16}\selectfont
\caption{Persistent contrastive divergence with buffer}\label{algo: PCD}
\begin{algorithmic}
\Require Discrete dataset $\mathcal{D}$. Integers $N_{\text{iters}}, N_{\text{batch}}, N_{\text{buffer}}$.
\Require Unnormalised log probability function $f(\cdot \ ; \ \bm{\theta})$. Step-size $\epsilon$. Optional regulariser $h(\bm{\theta})$.
\Require MCMC transition operator $\mathcal{T}( \cdot \given \cdot)$. Number of MCMC steps $K$.
\State $ \mathcal{B} \gets [\s^1, \ldots \s^{N_{\text{buffer}}} ]$ \Comment{Initialise buffer of persistent chains}
\For{$i \in \{1, \ldots, N_{\text{iters}} \}$}
\State Sample minibatch $B$ of size $N_{\text{batch}}$ from buffer $\mathcal{B}$
\For{$j \in \{1, \ldots, K \}$}
\State $B \sim \mathcal{T}( \cdot \given B)$ \Comment{Parallelised update to minibatch}
\EndFor
\State Update buffer $\mathcal{B}$ with the new values in $B$ \Comment{i.e.\ persist the chains}
\State Sample minibatch $X$ of size $N_{\text{batch}}$ from dataset $\mathcal{D}$
\State \vspace{-1.5em}\begin{flalign}
     \hspace{2em}\vg \ &\gets \ \frac{1}{N_{\text{batch}}} \ \Big[ \sum_{\s \in X} \nabla_{\bm{\theta}} f(\s; \ \bm{\theta}) \ - \ \sum_{\s \in B} \nabla_{\bm{\theta}} f(\s; \ \bm{\theta}) \Big] - \nabla_{\bm{\theta}} h(\bm{\theta}) & \\
     \bm{\theta} \ &\gets \ \bm{\theta} + \epsilon \vg
     \end{flalign}
\EndFor
\end{algorithmic}
}
\end{algorithm}

The gold-standard approach for the estimation of statistical models is maximum-likelihood estimation (MLE). Unfortunately, for unnormalised models
\begin{align}
    & \log p(\s; \ \bm{\theta}) = f(\s; \ \bm{\theta}) - \log Z(\bm{\theta}), & Z(\bm{\theta}) = \sum_{\s} \exp(f(\s; \ \bm{\theta}))
\end{align}
the normaliser $Z(\bm{\theta})$ is presumed intractable, and so we cannot actually compute $\log p(\s; \ \bm{\theta})$, which is required for MLE. However, we can conveniently express the gradient as
\begin{align}
    \nabla_{\bm{\theta}} \log p(\s ; \ \bm{\theta}) = \nabla_{\bm{\theta}} f_{\bm{\theta}}(\s) - \mathbb{E}_{p(\s; \ \bm{\theta})} [\nabla_{\bm{\theta}} f_{\bm{\theta}}(\s)]
\end{align}
and then use a Monte-Carlo estimate of the second term, where approximate samples are drawn from $p(\s; \ \bm{\theta})$ using an MCMC sampler. 

Running an MCMC sampler afresh every time we wish to perform a gradient-based parameter update quickly becomes prohibitively expensive. Persistent contrastive divergence provides a solution to this problem, by not starting afresh each time, but by \emph{persisting} MCMC chains across parameter updates. One version of PCD is given in Algorithm \ref{algo: PCD}. This implementation adopts vanilla stochastic gradient descent to update the model parameters, but alternative optimisers (e.g.\ Adam \citep{kingma2014adam}) can be substituted.

\section{Estimation of Ising models}
\label{appendix: ising lattice}
\begin{figure}[!t]
\centering
\begin{subfigure}{0.3\textwidth}
\includegraphics[width=.99\linewidth]{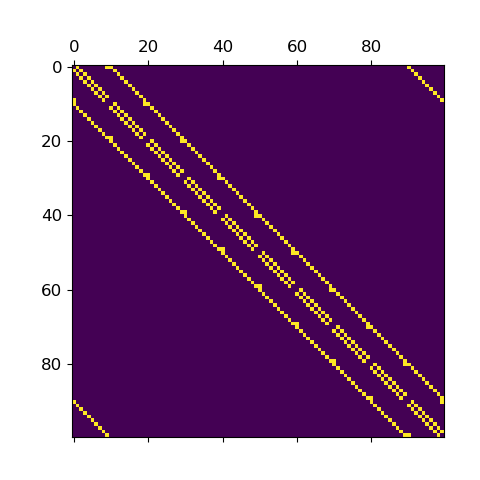}
\end{subfigure}\hspace{10em}
\begin{subfigure}{0.2\textwidth}
\includegraphics[width=.95\linewidth]{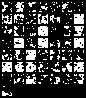}
\end{subfigure}
 \caption{Left: ground truth lattice Ising model with connection strength $\theta = 0.2$. Right: samples from this model generated with $1,000,000$ steps of Gibbs sampling.}
\label{fig: lattice ising matrix}
\end{figure}

The $100 \times 100$ Lattice Ising matrix used in our experiments was generated via \href{https://igraph.org/}{igraph}, specifically, we call \texttt{ig.Graph.Lattice(dim=[10, 10], circular=True)}, which returns a binary adjacency matrix, and then multiply this by a `connection strength' parameter $\theta = 0.2$. The resulting matrix, along with samples from the Ising model, is shown in Figure \ref{fig: lattice ising matrix}.

For learning, we use PCD as described in Algorithm \ref{algo: PCD}, replacing vanilla SGD with Adam. The dataset $\mathcal{D}$ consists of $10,000$ samples generated via $1,000,000$ steps of Gibbs sampling. We set $N_{\text{iters}}=2,000, N_{\text{batch}}=50, N_{\text{buffer}}=5000, \epsilon=0.0003$. Our model takes the form $f(\s ; \ J) = \s^T J \s$, where $J$ is unconstrained. We use the regulariser $h(J) = 0.01 \sum_{i, j} |J_{i, j}|$.

\section{Sampling convolutional energy-based models}
\label{appendix: conv ebm}
\begin{figure}[!b]
    \centering
    \begin{subfigure}{0.3\textwidth}
    \centering
    \begin{tabular}{l}
    \toprule
        \texttt{Conv2d}(1, 16, 3, 1, 1) \\
      \texttt{SiLU}() \\
      \midrule
        \texttt{Conv2d}(16, 32, 4, 2, 1) \\
      \texttt{SiLU}() \\
      \midrule
        \texttt{Conv2d}(32, 32, 3, 1, 1) \\
      \texttt{SiLU}() \\
      \midrule
        \texttt{Conv2d}(32, 64, 4, 2, 1) \\
      \texttt{SiLU}() \\
      \midrule
        \texttt{Conv2d}(64, 64, 3, 1, 1) \\
      \texttt{SiLU}() \\
      \midrule
        \texttt{Conv2d}(64, 128, 4, 2, 1) \\
      \texttt{SiLU}() \\
      \midrule
        \texttt{Conv2d}(128, 128, 2, 1, 0) \\
      \texttt{SiLU}() \\
    \bottomrule
  \end{tabular}
  \caption{7-layer convolutional neural network used as an EBM to model USPS digits. The \texttt{SiLU}() activation function \citep{ramachandran2017searching} is also called the `swish' activation.}
  \label{table: usps convnet architecture}
  \end{subfigure}\hspace{5em}
  \begin{subfigure}{0.45\textwidth}
    \begin{subfigure}{0.1\textwidth}
  \centering
    \includegraphics[scale=0.75]{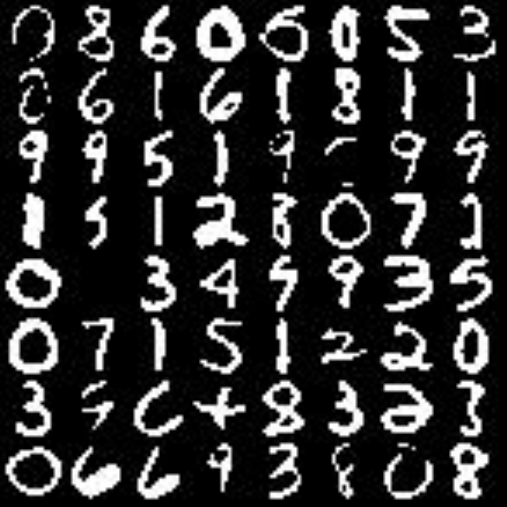}
  \end{subfigure}\hspace{8.3em}
   \begin{subfigure}{0.1\textwidth}
  \centering
    \includegraphics[scale=0.75]{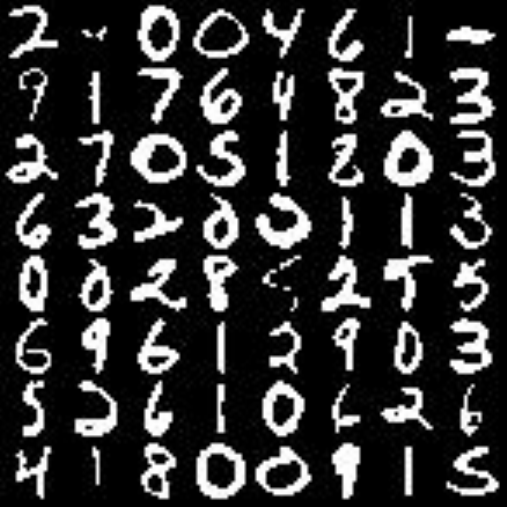}
  \end{subfigure} \\[1em]
   \begin{subfigure}{0.1\textwidth}
  \centering
    \includegraphics[scale=0.75]{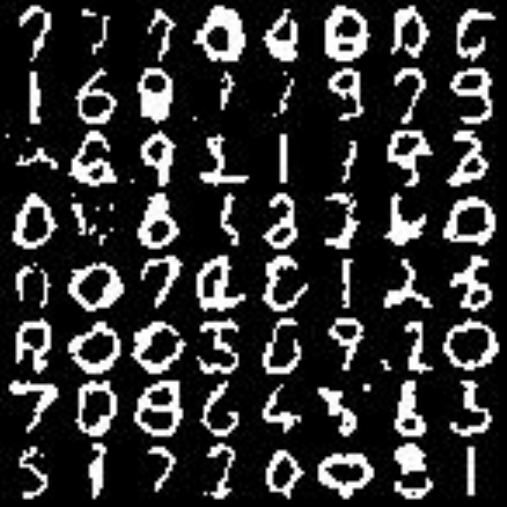}
  \end{subfigure} \hspace{8.em}
   \begin{subfigure}{0.1\textwidth}
  \centering
    \includegraphics[scale=0.75]{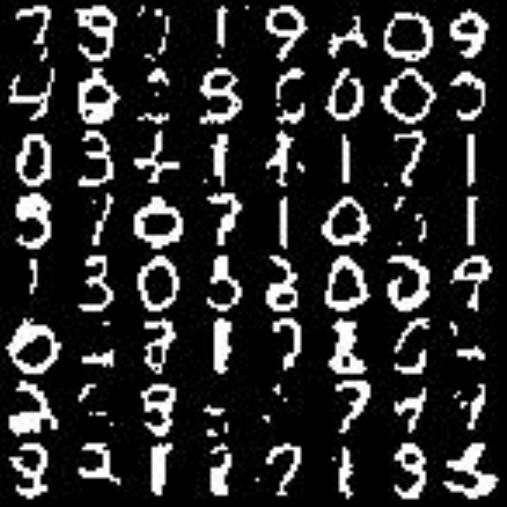}
  \end{subfigure}
    \caption{Top row: real images from the USPS dataset. \\ Bottom row: samples from ground-truth quadratic-EBM model trained with GWG and $K=50$.}
  \end{subfigure}
 \end{figure}

The architecture of the convolutional neural network used in our experiments is shown in Table \ref{table: usps convnet architecture}, alongside images from the USPS dataset. To learn a ground-truth quadratic-EBM model of the form $\frac{1}{2} \s^T J^* \s + f_{\theta^*}(\s)$, we use PCD as shown in Algorithm \ref{algo: PCD} with GWG as our sampler and $K=50$. We set $N_{\text{iters}}=10,000, N_{\text{batch}}=50, N_{\text{buffer}}=5,000, \epsilon=0.0003$. We use weight decay of $0.0001$ on the neural net weights.

After learning this ground-truth EBM, we then rerun PCD using the model $\frac{1}{2} \s^T J \s + f_{\theta^*}(\s)$, where $J$ is an unconstrained matrix of parameters. During this re-estimation phase, the settings of PCD remain the same, except for $N_{\text{iters}}=2,000$.

\begin{figure}
\centering
  \begin{subfigure}{0.1\textwidth}
  \centering
      {\Large \hphantom{FAKE}}
  \end{subfigure}
    \begin{subfigure}{0.2\textwidth}
  \centering
      {\Large K=5}
  \end{subfigure}
   \begin{subfigure}{0.2\textwidth}
  \centering
      {\Large K=10}
  \end{subfigure}
   \begin{subfigure}{0.2\textwidth}
  \centering
      {\Large K=15}
  \end{subfigure}
   \begin{subfigure}{0.2\textwidth}
  \centering
      {\Large K=20}
  \end{subfigure} \\[5ex]
    
  \begin{subfigure}{0.1\textwidth}
  \centering
    {\Large NCG}
  \end{subfigure}
    \begin{subfigure}{0.2\textwidth}
  \centering
    \vspace{1em} \caption*{0.128} \vspace{-0.5em}
    \includegraphics[scale=0.6]{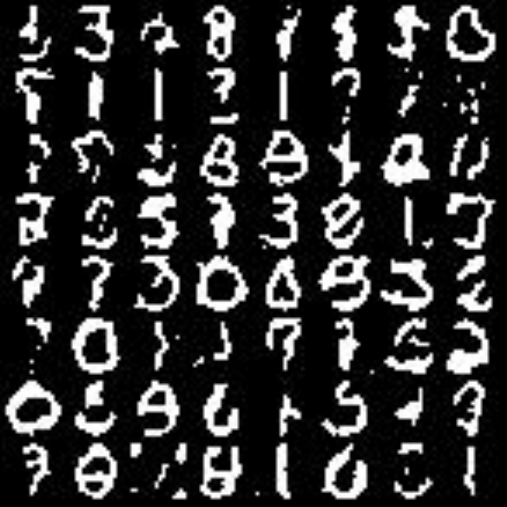}
  \end{subfigure}\hspace{1em}
   \begin{subfigure}{0.2\textwidth}
  \centering
    \vspace{1em}   \caption*{0.116} \vspace{-0.5em}
    \includegraphics[scale=0.6]{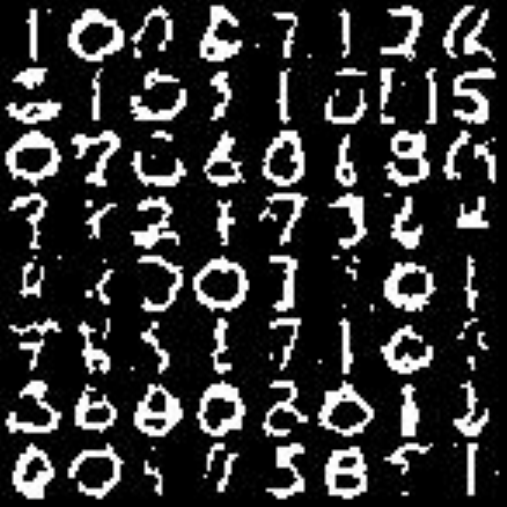}
  \end{subfigure}\hspace{1em}
   \begin{subfigure}{0.2\textwidth}
  \centering
    \vspace{1em}   \caption*{0.112} \vspace{-0.5em}
    \includegraphics[scale=0.6]{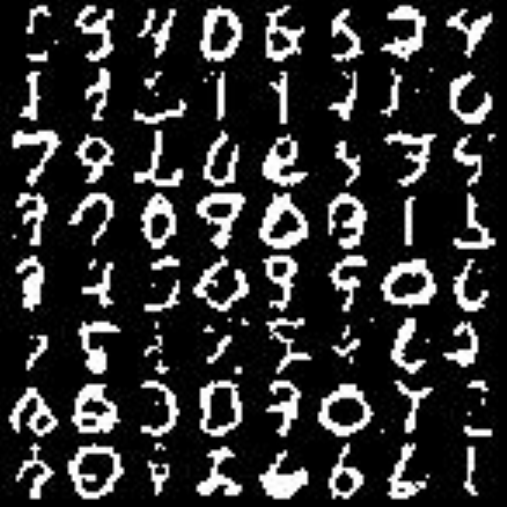}
  \end{subfigure}\hspace{1em}
   \begin{subfigure}{0.2\textwidth}
  \centering
    \vspace{1em}   \caption*{0.112} \vspace{-0.5em}
    \includegraphics[scale=0.6]{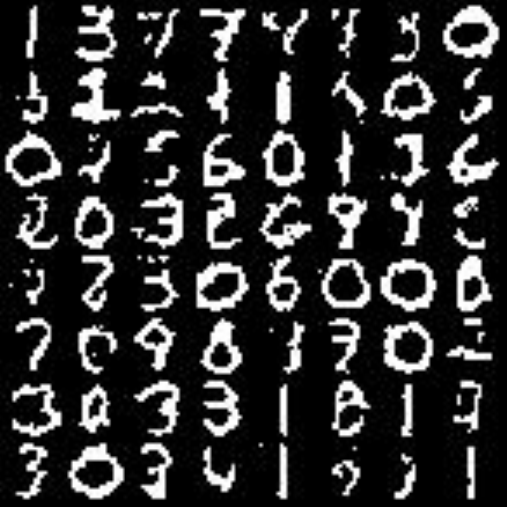}
  \end{subfigure} \\
    
  \begin{subfigure}{0.1\textwidth}
  \centering
    {\Large AVG}
  \end{subfigure}
    \begin{subfigure}{0.2\textwidth}
  \centering
    \vspace{1em}   \caption*{3.310} \vspace{-0.5em}
    \includegraphics[scale=0.6]{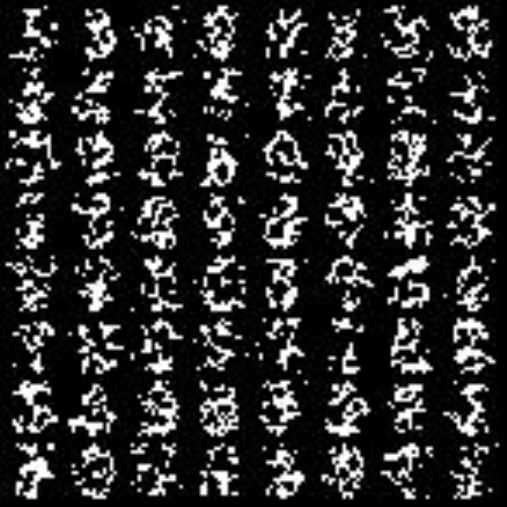}
  \end{subfigure}\hspace{1em}
   \begin{subfigure}{0.2\textwidth}
  \centering
    \vspace{1em}   \caption*{2.599} \vspace{-0.5em}
    \includegraphics[scale=0.6]{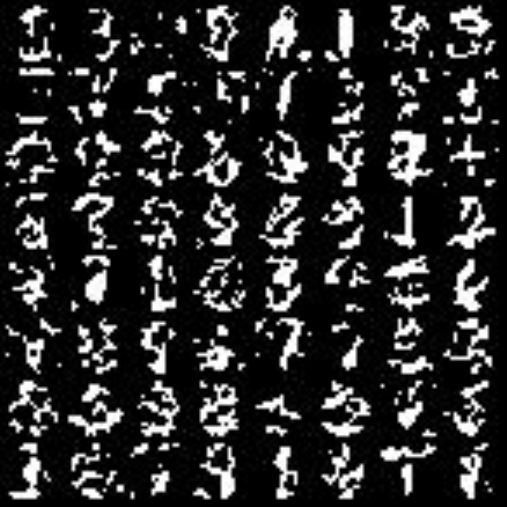}
  \end{subfigure}\hspace{1em}
   \begin{subfigure}{0.2\textwidth}
  \centering
    \vspace{1em}  \caption*{0.496} \vspace{-0.5em}
    \includegraphics[scale=0.6]{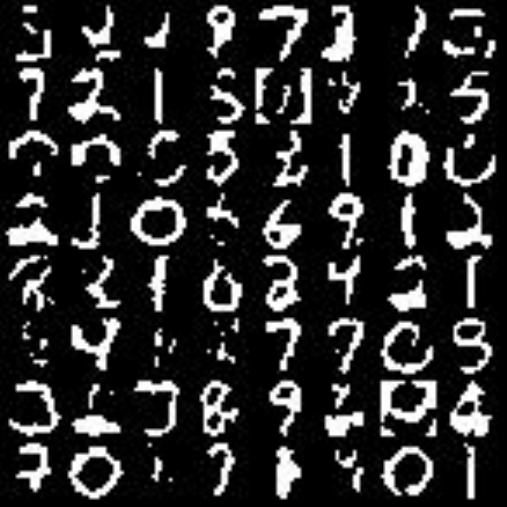}
  \end{subfigure}\hspace{1em}
   \begin{subfigure}{0.2\textwidth}
  \centering
    \vspace{1em}   \caption*{0.114} \vspace{-0.5em}
    \includegraphics[scale=0.6]{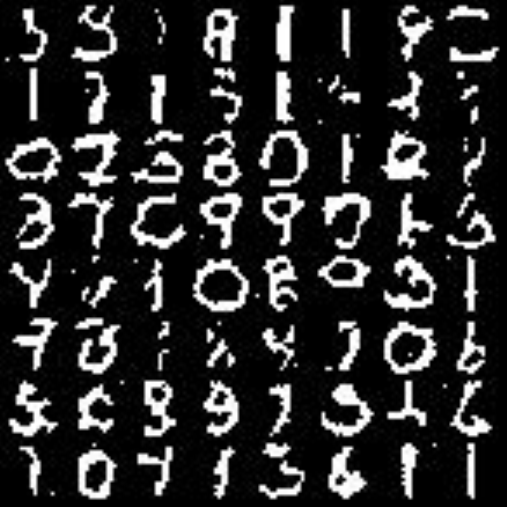}
  \end{subfigure} \\
 
  \begin{subfigure}{0.1\textwidth}
  \centering
    {\Large PAVG}
  \end{subfigure}
    \begin{subfigure}{0.2\textwidth}
  \centering
    \vspace{1em}   \caption*{3.212} \vspace{-0.5em}
    \includegraphics[scale=0.6]{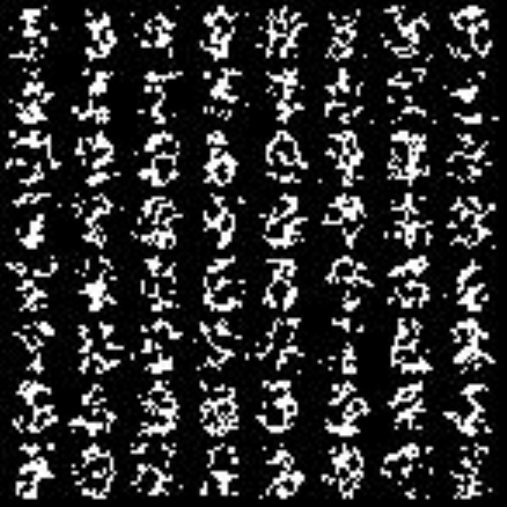}
  \end{subfigure}\hspace{1em}
   \begin{subfigure}{0.2\textwidth}
  \centering
    \vspace{1em}   \caption*{0.117} \vspace{-0.5em}
    \includegraphics[scale=0.6]{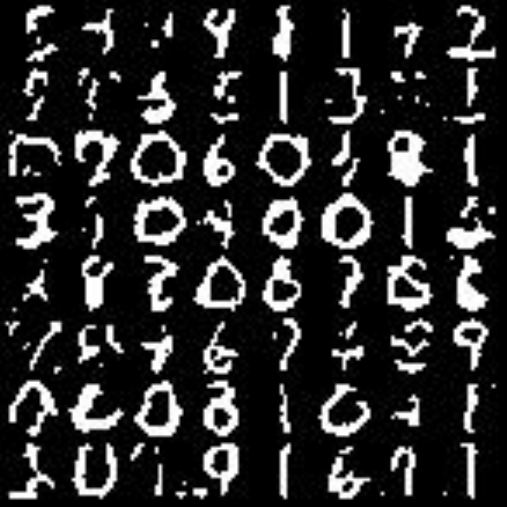}
  \end{subfigure}\hspace{1em}
   \begin{subfigure}{0.2\textwidth}
  \centering
    \vspace{1em}  \caption*{0.114} \vspace{-0.5em}
    \includegraphics[scale=0.6]{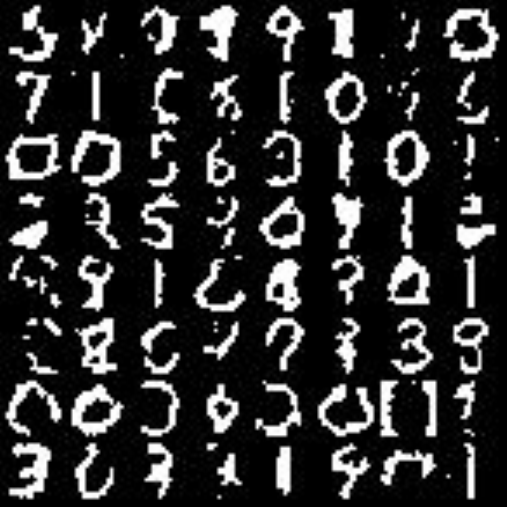}
  \end{subfigure}\hspace{1em}
   \begin{subfigure}{0.2\textwidth}
  \centering
    \vspace{1em}   \caption*{0.112} \vspace{-0.5em}
    \includegraphics[scale=0.6]{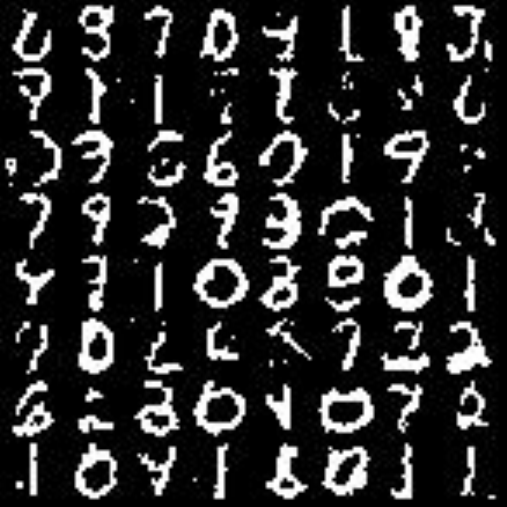}
  \end{subfigure} \\

  \begin{subfigure}{0.1\textwidth}
  \centering
    {\Large GWG}
  \end{subfigure}
    \begin{subfigure}{0.2\textwidth}
  \centering
    \vspace{1em}  \caption*{0.732} \vspace{-0.5em}
    \includegraphics[scale=0.6]{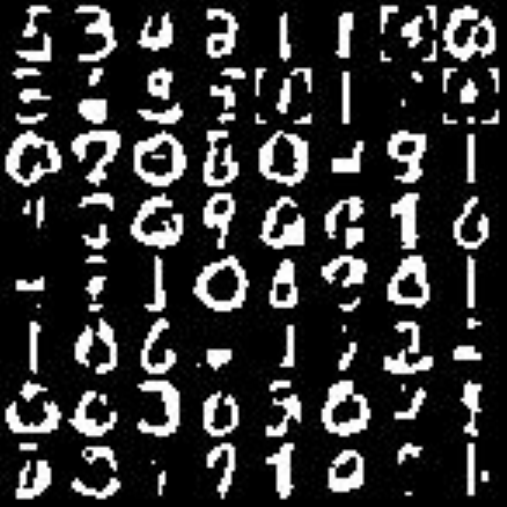}
  \end{subfigure}\hspace{1em}
   \begin{subfigure}{0.2\textwidth}
  \centering
    \vspace{1em}   \caption*{0.156} \vspace{-0.5em}
    \includegraphics[scale=0.6]{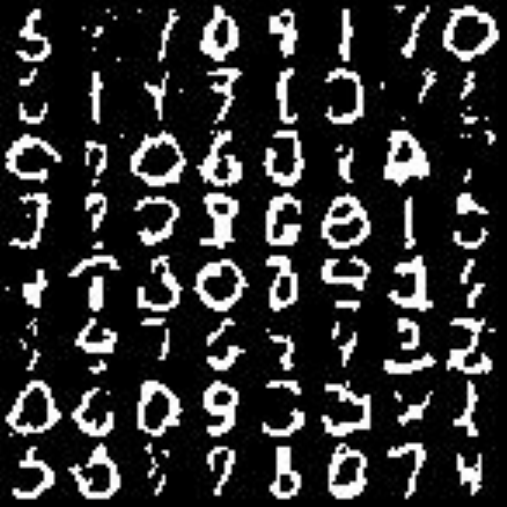}
  \end{subfigure}\hspace{1em}
   \begin{subfigure}{0.2\textwidth}
  \centering
    \vspace{1em}   \caption*{0.120} \vspace{-0.5em}
    \includegraphics[scale=0.6]{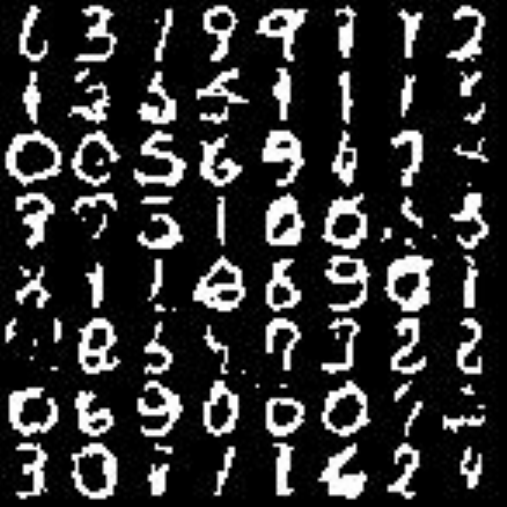}
  \end{subfigure}\hspace{1em}
   \begin{subfigure}{0.2\textwidth}
  \centering
    \vspace{1em}  \caption*{0.114} \vspace{-0.5em}
    \includegraphics[scale=0.6]{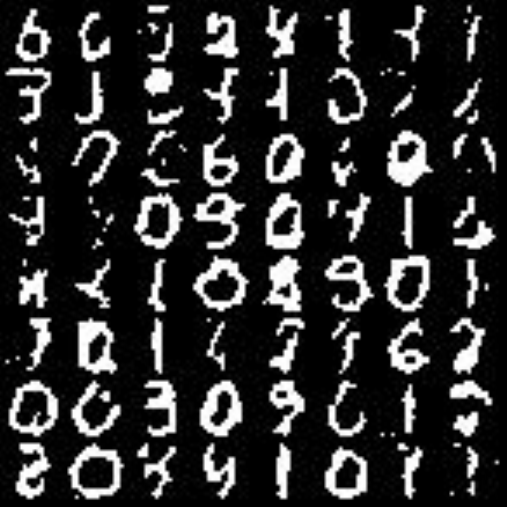}
  \end{subfigure} \\

  \begin{subfigure}{0.1\textwidth}
  \centering
    {\Large Gibbs}
  \end{subfigure}
    \begin{subfigure}{0.2\textwidth}
  \centering
  \vspace{1em}   \caption*{3.121} \vspace{-0.5em}
    \includegraphics[scale=0.6]{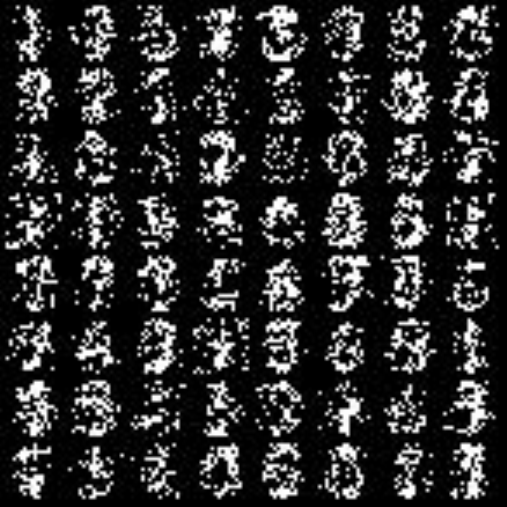}
  \end{subfigure}\hspace{1em}
   \begin{subfigure}{0.2\textwidth}
  \centering
    \vspace{1em} \caption*{2.270} \vspace{-0.5em}
    \includegraphics[scale=0.6]{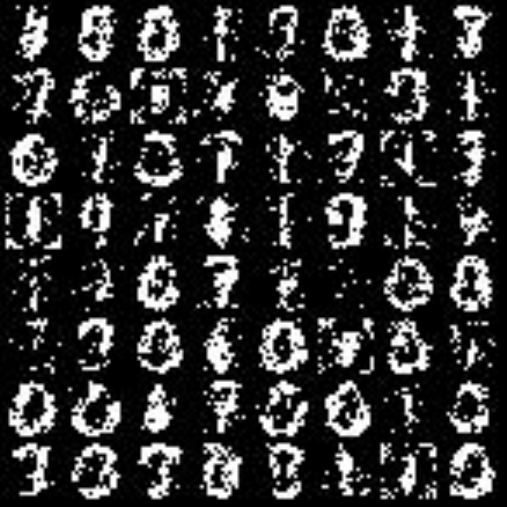}
  \end{subfigure}\hspace{1em}
   \begin{subfigure}{0.2\textwidth}
  \centering
    \vspace{1em} \caption*{1.637} \vspace{-0.5em}
    \includegraphics[scale=0.6]{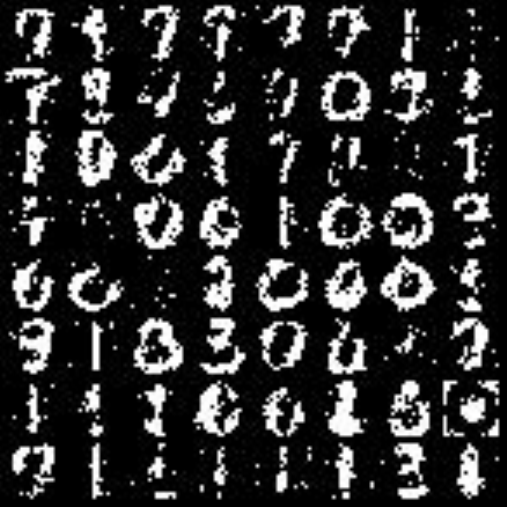}
  \end{subfigure}\hspace{1em}
   \begin{subfigure}{0.2\textwidth}
  \centering
  \vspace{1em} \caption*{1.204} \vspace{-0.5em}
    \includegraphics[scale=0.6]{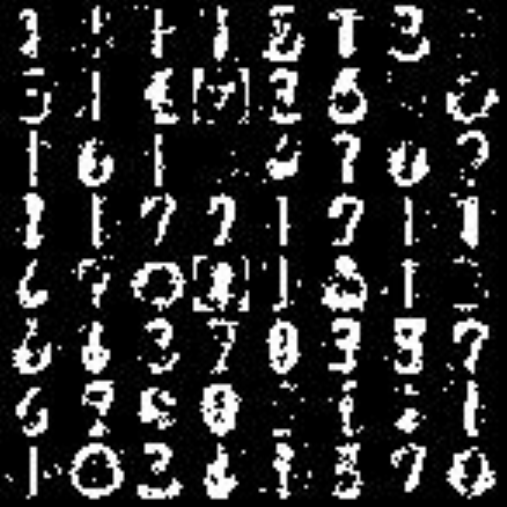}
  \end{subfigure} \\
  \caption{Samples for the methods described in Table \ref{table: conv table}. Above each image is the corresponding estimation error.}
  \label{fig: mcmc buffers per method for USPS quadratic-EBM}
\end{figure}

\end{document}